\pdfoutput=1

\documentclass[11pt]{article}

\usepackage{ACL2023}

\usepackage{times}
\usepackage{latexsym}

\usepackage[T1]{fontenc}

\usepackage[utf8]{inputenc}

\usepackage{microtype}

\usepackage{inconsolata}

\usepackage{epsfig}
\usepackage{graphicx}
\usepackage{amsmath}
\usepackage{amssymb}
\usepackage{booktabs}

\usepackage{subcaption}
\usepackage{multicol}
\usepackage{multirow}
\usepackage{bbding}

\newcommand\und{\underline}

\newcommand{\Houlsby}{\textsc{Houlsby}}
\newcommand{\Pfeiffer}{\textsc{Pfeiffer}}
\newcommand{\Parallel}{\textsc{Parallel}}
\newcommand{\Lora}{\textsc{LoRa}}
\newcommand{\Compacter}{\textsc{Compacter}}
\newcommand{\Prefix}{\textsc{Prefix}}
\newcommand{\Inverse}{\textsc{Inverse}}

\newcommand{\RQ}[1]{\textsc{\bf RQ#1}}

\DeclareMathAlphabet{\mathsfit}{\encodingdefault}{\sfdefault}{m}{sl}
\SetMathAlphabet{\mathsfit}{bold}{\encodingdefault}{\sfdefault}{bx}{n}

\DeclareMathAlphabet{\pazocal}{OMS}{zplm}{m}{n}

\usepackage{makecell}
\usepackage{bbding}

\title{Towards Parameter-Efficient Integration of Pre-Trained Language Models In Temporal Video Grounding}

\author{Erica K. Shimomoto\textsuperscript{1}, Edison Marrese-Taylor\textsuperscript{1}, Hiroya Takamura\textsuperscript{1}, \\
\bf Ichiro Kobayashi\textsuperscript{1,2}, \bf Hideki Nakayama\textsuperscript{1,3}, \bf Yusuke Miyao\textsuperscript{1,3} \\
National Institute of Advanced Industrial Science and Technology\textsuperscript{1}\\
Ochanomizu University\textsuperscript{2}, The University of Tokyo\textsuperscript{3}\\
{\tt \{kidoshimomoto.e,edison.marrese,takamura.hiroya\}@aist.go.jp}\\
{\tt koba@is.ocha.ac.jp, nakayama@ci.i.u-tokyo.ac.jp, yusuke@is.s.u-tokyo.ac.jp}\\}

\begin{document}
\maketitle
\begin{abstract}

This paper explores the task of Temporal Video Grounding (TVG) where, given an untrimmed video and a natural language sentence query, the goal is to recognize and determine temporal boundaries of action instances in the video described by the query. 
Recent works tackled this task by improving query inputs with large pre-trained language models (PLM) at the cost of more expensive training. However, the effects of this integration are unclear, as these works also propose improvements in the visual inputs.  
Therefore,  this paper studies the effects of PLMs in TVG and assesses the applicability of parameter-efficient training with NLP adapters. 
We couple popular PLMs with a selection of existing approaches and test different adapters to reduce the impact of the additional parameters. 
Our results on three challenging datasets show that, without changing the visual inputs, TVG models greatly benefited from the PLM integration and fine-tuning, stressing the importance of sentence query representation in this task.
Furthermore, NLP adapters were an effective alternative to full fine-tuning, even though they were not tailored to our task, allowing PLM integration in larger TVG models and delivering results comparable to SOTA models. 
Finally, our results shed light on which adapters work best in different scenarios.

\end{abstract}
\section{Introduction}

Temporal Video Grounding (TVG) is a fundamental task in Computer Vision (CV), where the goal is to have models recognize and determine temporal boundaries of action instances in videos \cite{Shou_2016_CVPR,guAVAVideoDataset2018,girdharVideoActionTransformer2019} using queries provided in natural language \cite{Gao_2017_ICCV,Hendricks_2017_ICCV}. 
Over the past few years, interest in this task has grown substantially due to its complexity and potential applications, which has led to the release of several models that either use propose-and-rank techniques or directly predict the starting and ending temporal locations \cite{ghosh_excl_2019,rodriguez_proposal-free_2020}. 

Following trends in other vision-and-language (V\&L) tasks, some of the latest models combine vision encoders and pre-trained language models (PLM). 
We find works that directly encode the query using PLMs \cite{nawazTemporalMomentLocalization2022,wangNegativeSampleMatters2022}, which are later fine-tuned with the rest of the architecture, or that try to project both the query and video into the same embedding space using a Transformer \cite{zhangMultiStageAggregatedTransformer2021}.

Despite the performance improvements, it is difficult to isolate the effects of the improved language representations, as these works also propose new video-language matching approaches or use different video encoders. 
Another drawback is their computational cost for training, as parameter counts grow substantially once PLMs are incorporated. 

To address this problem, several parameter-efficient training methods have been recently proposed for both Natural Language Processing (NLP)  \cite{karimimahabadiParameterefficientMultitaskfinetuning2021} and CV models \cite{rebuffiLearningMultipleVisual2017}. 
Among these approaches, adapters \cite{houlsbyParameterEfficientTransferLearning2019,bapnaSimpleScalableAdaptation2019} and their variations have been particularly effective\footnote{While the term ``adapter" is often used to refer to the original adapter proposed by \citet{houlsbyParameterEfficientTransferLearning2019}, we use it to refer to any efficient fine-tuning method in this work.}, as they lead to performance as high as fine-tuning while training only a small set of parameters.
Adapters have been successfully combined with vision models for several tasks \cite{kimHowAdaptYour2021,zhouLearningPromptVisionLanguage2022}, showing that using a few parameters to learn to fuse vision and language representations without losing performance is possible.
However, we note that efforts so far have focused only on image \cite{zhangTipAdapterTrainingfreeCLIPAdapter2021} and video \cite{panSTAdapterParameterEfficientImagetoVideo2022} classification tasks or on leveraging pre-trained generative models, e.g., by re-casting existing vision-and-language tasks as language generation \cite{sungVLAdapterParameterEfficientTransfer2022}. In contrast, as the training signal in TVG comes from the visual modality, we cannot cast it as language generation.

Therefore, this paper studies the effects of large PLMs in the TVG task and investigates the applicability of NLP adapters for a parameter-efficient integration.
We couple popular PLM models with a selection of previous works, allowing us to isolate and understand their effects on performance.
Concretely, we analyze ExCL \cite{ghosh_excl_2019}, TMLGA~\cite{rodriguez_proposal-free_2020}  and DORi \cite{rodriguez-opazo_dori_2021}, three proposal-free TVG models with different levels of complexity. 
Moreover, we also benchmark several parameter-efficient training alternatives based on adapters.

We conduct thorough experiments on three challenging datasets, Charades-STA \cite{Gao_2017_ICCV}, ActivityNet Captions \cite{krishnaDenseCaptioningEventsVideos2017} and YouCookII \cite{zhouAutomaticLearningProcedures2018,zhouWeaklySupervisedVideoObject2018}, covering videos and queries with varying lengths of different activities. 
Concretely, we seek to answer the following research questions:  \RQ{1}: Does incorporating a PLM improve the performance of existing TVG models?; \RQ{2}: Are adapters an alternative to full fine-tuning of the PLM parameters within TVG?; \RQ{3}: Is there an adapter that works best for TVG?; \RQ{4}: What is the impact of different PLMs?; \RQ{5}: How does the combination of existing TVG models with PLMs trained with adapters perform against state-of-the-art models?

Our results offer concrete answers to these questions, helping us clarify the role of PLMs in the TVG task and quantify how much they can improve the existing model's grounding capabilities. They suggest that, by only changing the query sentence representation using PLMs, TVG models can greatly improve performance, especially when PLMs are fine-tuned, stressing the importance of the text query representation in this task.

Our contributions can be summarized as follows: (1) We quantify the impact of PLMs in TVG models, (2) We perform the first work on benchmarking different types of adapters on the TVG task, shedding light on which adapters work best for each case, and (3) We offer an empirical demonstration that adapters can reach or surpass the performance of full fine-tuning while updating only $\sim10\%$ of the parameters in our task. The code to reproduce our experiments is available at \url{github.com/ericashimomoto/parameter-efficient-tvg}.

\section{Related Work}
\noindent \textbf{Temporal Video Grounding:} 
Work on this task can be divided into two main approaches. On the one hand, we find techniques based on proposal generation, where the idea is to, given a query, output a set of candidate clips which could later be ranked \cite{liu2018attentive, ge2019mac}. Further research has mostly focused on reducing the number of proposals by producing query-guided or query-dependent approaches \cite{chen-etal-2018-temporally,sap2019,xu2019multilevel}, or on creating maps that can cover diverse video moments with different lengths \cite{zhangMultiScale2DTemporal2021}. \citet{zhangMultiStageAggregatedTransformer2021} adopted a Transformer-based multi-modal model (MSAT) which is pre-trained for this setting. More recently, models have incorporated contrastive losses to improve performance further. This is the case of both CPL \cite{zhengWeaklySupervisedTemporal2022} and MNM \cite{wangNegativeSampleMatters2022}, which also incorporate Transformer-based components in their pipelines.

The second line of approaches has instead proposed to directly predict the start and end locations through the video span \citep[ExCL;][]{ghosh_excl_2019}. 
Work on this line of research has focused on improving the performance by modelling label uncertainty \citep[TMLGA;][]{rodriguez_proposal-free_2020}, adding spatial features \citep[DORi;][]{rodriguez-opazo_dori_2021} or improving the text-to-video matching strategies \cite{munLocalGlobalVideoTextInteractions2020,zengDenseRegressionNetwork2020}. For example, CPN \cite{zhaoCascadedPredictionNetwork2021} adopted an ad-hoc graph-based technique. Other approaches have focused on exploiting local and global features for better performance like \citet{munLocalGlobalVideoTextInteractions2020}. CPNet \cite{liProposalFreeVideoGrounding2021} recently proposed a pyramid-like approach where the model progressively replenishes the temporal contexts and refines the location of the queried activity by enlarging the temporal receptive fields. Finally, two recent approaches, VSLNet \cite{zhangSpanbasedLocalizingNetwork2020} and BCPN, \cite{nawazTemporalMomentLocalization2022} have proposed to cast the task as visual question answering.

On top of these lines, recently, an effort has been made to solve a variant of the task named spatio-temporal video grounding. In this case, besides predicting when the moment starts and ends, the model should also identify where the action described by the textual query occurs in the frames. Works on this task heavily rely on the transformers architecture~\cite{yang2022tubedetr}, with significant effort in eliminating the need for any pre-trained object detectors~\cite{su2021stvgbert, jin2022embracing}. While closely related to our task, due to the addition of the spatial dimension, this task naturally emphasizes the role of the visual modality in the grounding, further deviating from our language-driven approach. Therefore, our study does not consider models tailored for this task.

\begin{figure}[t]
        \centering
        \includegraphics[width=\linewidth]{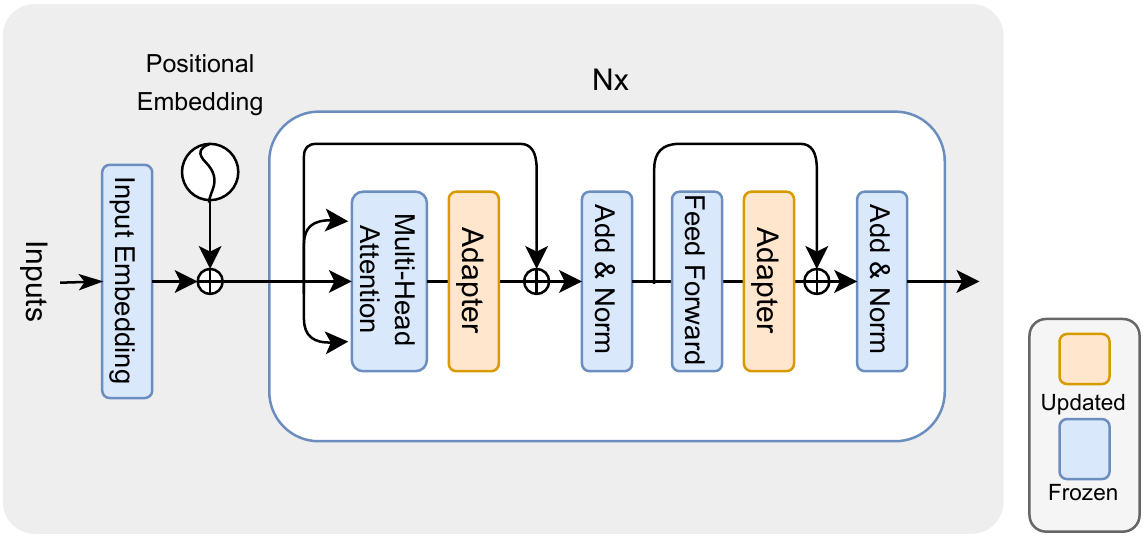}
        \caption{Illustration of the traditional bottleneck adapter proposed by \citet{houlsbyParameterEfficientTransferLearning2019}. The adapter layers are introduced after the multi-head attention and feed-forward layers. Orange color refers to trainable parameters, and blue color refers to frozen ones.}
        \label{fig:adapter}
        \vspace{-4mm}
\end{figure}

\noindent \textbf{Parameter-efficient model training:} As machine learning models continue to grow, updating their parameters efficiently is becoming increasingly important. One key idea has been to only update newly-added parameters \cite{rebuffiLearningMultipleVisual2017,rebuffiEfficientParametrizationMultiDomain2018}. With the advent of large pre-trained Transformer-based models in NLP, 
this idea has led to the development of adapters \cite{houlsbyParameterEfficientTransferLearning2019}: sub-networks with few parameters that are inserted after every attention and feed-forward layer in a given model, as illustrated in Figure~\ref{fig:adapter}. We also find a variety of prompt-based approaches, which add trainable parameters into the model inputs \cite{liPrefixTuningOptimizingContinuous2021,lesterPowerScaleParameterEfficient2021,guPPTPretrainedPrompt2022}. Alternative techniques, such as sparsely updating a small number of parameters of the model \cite{benzakenBitFitSimpleParameterefficient2022,guoParameterEfficientTransferLearning2021,sungTrainingNeuralNetworks2021}, or low-rank factorization for the weights to be updated \cite{mahabadi2021parameter,karimimahabadiCompacterEfficientLowRank2021,huLoRALowRankAdaptation2022} have also been proposed recently. Finally, \citet{heUnifiedViewParameterEfficient2022,maoUniPELTUnifiedFramework2022} combined some of these techniques to propose a unified parameter-efficient training framework. Though we focus on NLP adapters, we deviate from previous work as our approach incorporates the visual modality.

Only updating newly-added parameters has also been proposed in CV, with some work predating the advent of Transformers \cite{rebuffiLearningMultipleVisual2017,rebuffiEfficientParametrizationMultiDomain2018}. More recently, we find works combining pre-trained language models with multi-modal inputs. For example, \citet{tsimpoukelliMultimodalFewShotLearning2021} trained a vision encoder to represent each image as a sequence of continuous embeddings, such that a frozen language model prompted with this prefix generates the appropriate image caption. \citet{yangZeroShotVideoQuestion2022} showed that it is possible to perform zero-shot video question answering by leveraging frozen bidirectional language models. More recently, \citet{sungVLAdapterParameterEfficientTransfer2022} cast multiple V\&L tasks as text generation, combining NLP adapters with pre-trained encoder-decoders, such as BART \cite{lewisBARTDenoisingSequencetoSequence2020}, with existing image encoders, such as CLIP \cite{radfordLearningTransferableVisual2021}. The latter has also lately been the target of several studies that extend parameter-efficient techniques for CV \cite{kimHowAdaptYour2021,zhangTipAdapterTrainingfreeCLIPAdapter2021,zhouLearningPromptVisionLanguage2022}. 

Though our approach follows a similar trend, our interest lies in a grounding task where the training signal comes from the visual modality, which keeps us from casting our task as language generation.

\section{Proposed Approach}

\begin{figure}[t]
    \centering
   \includegraphics[width=\linewidth]{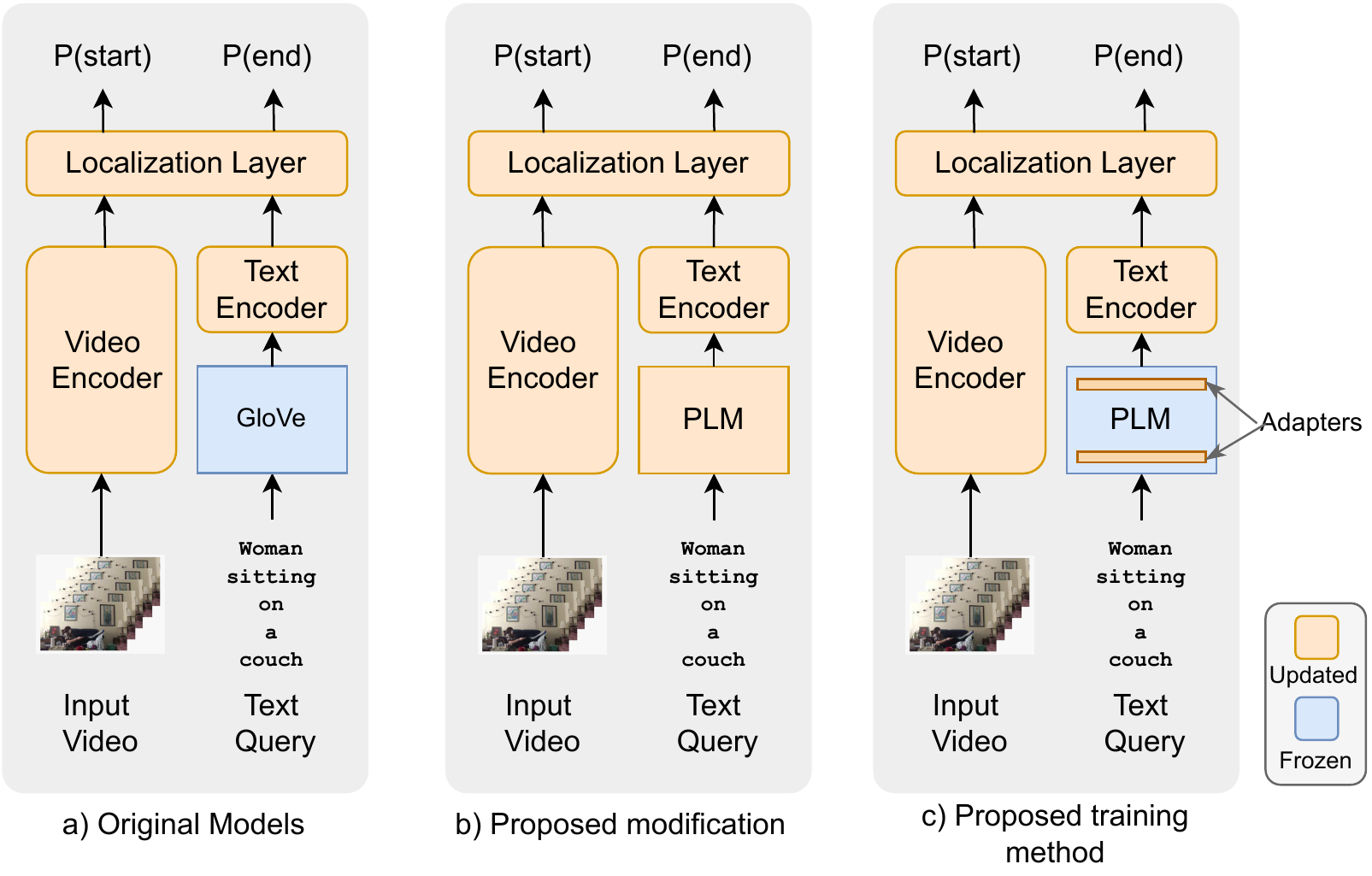}
   \caption{Illustration of our proposed approach to incorporate pre-trained language encoders and adapters into existing Temporal Video Grounding pipelines. Orange color refers to trainable parameters, and blue color refers to frozen ones.}
    \label{fig:proposed}
    \vspace{-4mm}
\end{figure}

Consider a video $V \in \pazocal{V}$, represented as a sequence of frames such that $V = \{v_{t}\}$ with $t=1, \dots, l$. Each video in $\pazocal{V}$ is annotated with a natural language passage $S \in \pazocal{S}$, where $S$ is a sequence of words $S = \{s_j\}$ with $j=1, \dots, m$, which describes what is happening at a certain period of time in the video. This interval is formally defined by $t_s$ and $t_e$, the starting and ending points of the annotations in time, respectively. The goal of the temporal video grounding task is to predict $t_s$ and $t_e$ given the video $V$ and the text query $S$.

\subsection{Temporal Video Grounding Models}
For this study, we focused on proposal-free models, which generally offer better performance, and we were careful only to consider works that used word embeddings. Furthermore, we only considered models which have their implementation available. Concretely, we selected ExCL, TMLGA, and DORi. ExCL was the first proposal-free model for TVG; TMLGA improved on it by handling video annotation uncertainty, and later DORi incorporated spatial features.

As shown in Figure~\ref{fig:proposed} (a), these models can be summarized into three main parts: a sentence encoder, a video encoder, and a localization module, which combines information from both modalities and predicts the start and end points of the segment in the video described by the query. While these models heavily explore information from the video input, they make relatively simple use of the sentence query, mainly processing pre-trained word embeddings, such as GloVe~\cite{pennington2014glove}, through a recurrent neural network. 

To better understand the role of improved language representations on the performance in the TVG task, we study the effects of PLMs by incorporating them as-is into a selection of existing models from the literature, therefore effectively isolating their impact on the performance across a range of settings. 
We tested several ways to incorporate PLMs, such as entirely replacing the text encoder block with the PLM, where most failed to deliver any performance improvement, and, therefore, decided to replace only the word embeddings, as shown in Figure~\ref{fig:proposed} (b). 

\subsection{Adapters}
\label{sec:adapters}

A concern when integrating large PLMs with existing TVG models is the alarming number of parameters added. While some TVG models are pretty small, recent models are exponentially increasing in size. For example, DORi has about 10M parameters, more than double than TMLGA (about 4M). Combining these models even with reasonably-sized PLMs, such as BERT \cite{devlinBERTPretrainingDeep2019}, increases the number of trainable parameters to over 120M. Also, the more sophisticated visual features are used, the larger the training data becomes. For example, the visual features for TMLGA are about 2.1MB per video in the ActivityNet dataset, while for DORi, they are about 82MB.  

To alleviate this issue, we also investigate several parameter-efficient training alternatives based on adapters to incorporate these PLMs into the existing model pipelines with reduced computational cost, as shown in Figure~\ref{fig:proposed} (c).




For our experiments, we adopt a large selection of adapters following previous work \cite{sungVLAdapterParameterEfficientTransfer2022}, including bottleneck adapters, such as the ones proposed by \citet{houlsbyParameterEfficientTransferLearning2019} (\Houlsby), \citet{pfeifferMADXAdapterBasedFramework2020} (\Pfeiffer{}), and \citet{heUnifiedViewParameterEfficient2022} (\Parallel); Invertible adapters (\Inverse) \cite{pfeifferMADXAdapterBasedFramework2020}, Prefix Tuning (\Prefix) \cite{liPrefixTuningOptimizingContinuous2021}, Compacter (\Compacter) \cite{karimimahabadiCompacterEfficientLowRank2021}, and LoRA (\Lora) \cite{huLoRALowRankAdaptation2022}. 

We note that some of the adapters we consider in our study were designed for specific purposes in NLP.
Our decision to still include such adapters in our experimental framework is motivated by their relative success in other vision-and-language tasks \cite{heParameterefficientFinetuningVision2022,kimHowAdaptYour2021,sungVLAdapterParameterEfficientTransfer2022}. 
Moreover, as our approach differs from existing work in this context, we were interested in offering empirical evidence to further understand these adapters' role in multi-modal scenarios.

\section{Experimental Framework}

\subsection{Datasets}

\textbf{Charades-STA:} Built upon the Charades dataset \cite{sigurdssonHollywoodHomesCrowdsourcing2016a}, it provides time-based annotations using a pre-defined set of activity classes and general video descriptions. We use the pre-defined train and test sets containing 12,408 and 3,720 moment-query pairs, respectively. Videos are 31s long on average and have a maximum duration of 194s, with 2.4 moments on average, each being 8.2s long on average and described using 7.2 words on average. 

\textbf{ActivityNet Captions:} Introduced by \citet{krishnaDenseCaptioningEventsVideos2017} and initially constructed for dense video captioning, it consists of 20k YouTube videos with an average length of 120s and a maximum duration of 755s. The videos contain 3.65 temporally localized time intervals and sentence descriptions on average, where descriptions have on average 13.48 words. Following previous works, we report the performance on the combined validation sets.

\textbf{YouCookII:} It consists of 2,000 long untrimmed videos from 89 cooking recipes obtained from YouTube by \citet{zhouAutomaticLearningProcedures2018,zhouWeaklySupervisedVideoObject2018}. Each step for cooking these dishes was annotated with temporal boundaries and aligned with the corresponding section of the recipe. The average video length is 316s and a maximum duration of 755s. Regarding relevant moment segments, each video has 7.73 moments on average, with each segment being 19.63s long and described using 8.7 words on average.

\subsection{Implementation Details}
\label{section:implementation_details}

\textbf{Temporal Video Grounding Models:} Our implementation for TMLGA and DORi is built on top of the original code released by the authors, while we use our own implementation of ExCL. To represent the video input, we follow the original implementations as close as possible. For TMLGA and DORi, we use the I3D features released with the respective papers. 
For ExCL, we use the features released by the DORi paper, extracted at 25 fps instead of the original 5 fps\footnote{In practice, this difference in fps is not an issue, as \citet{rodriguez-opazo_dori_2021} has shown that extraction at 25fps leads to better performance}. Specifically for DORi, we also use the spatial features released with the paper. 

\textbf{Pre-trained Language Models:} We combine our selected models  with pre-trained BERT, RoBERTa \cite{liuRoBERTaRobustlyOptimized2019} and DeBERTa \cite{heDeBERTaDecodingEnhancedBERT2022}, with the implementations provided by the HuggingFace library \cite{wolf2020transformers}. Furthermore, we used ``bert-base-uncased", ``roberta-base", and ``deberta-base" pre-trained models.

\textbf{Adapters:} We use the implementation provided by the adapter-transformers library \cite{pfeiffer2020adapterhub}, with default configurations. In particular, for the Invertible adapters (\Inverse), we only tested the ``PfeifferInvConfig", the original configuration proposed in the paper~\cite{pfeifferMADXAdapterBasedFramework2020}.

\textbf{Training:} Our experiments were performed on a 40-GB NVIDIA A100 GPU. Models were trained using ADAM \cite{kingmaAdamMethodStochastic2020} with a step-based learning rate scheduler. When fine-tuning the PLMs, we used a scheduler with linear warmup for the PLM parameters, while keeping the learning rate fixed for the rest of the parameters. 
For more details on hyper-parameters, we refer the readers to the Appendices~\ref{sec:app:detailedBERT} and \ref{sec:app:detailed_rob_deb}. 
The evaluation follows previous work, based on two widely used metrics \cite{Gao_2017_ICCV}, namely the Recall at various thresholds of the temporal Intersection over Union (tIoU or $R@\alpha$) measuring the percentage of predictions that have tIoU with ground truth larger than certain $\alpha$, and the mean averaged tIoU (mIoU). We use $\alpha$ threshold values of $0.3$, $0.5$ and $0.7$.

\section{Results and Discussion}

\begin{table*}[t]
\centering
    \scalebox{0.8}{
    \begin{tabular}{l@{\hspace{0.15cm}} c@{\hspace{0.15cm}} c@{\hspace{0.15cm}} c@{\hspace{0.15cm}} c@{\hspace{0.15cm}} c@{\hspace{0.15cm}} c@{\hspace{0.15cm}} c@{\hspace{0.15cm}} c@{\hspace{0.15cm}} c@{\hspace{0.15cm}} c@{\hspace{0.15cm}} c@{\hspace{0.15cm}} c@{\hspace{0.15cm}} c@{\hspace{0.15cm}}}
        \toprule
        \multirow{2}{*}{\textbf{Method}} & \multirow{2}{*}{\textbf{Params.}} & \multicolumn{4}{c}{\textbf{Charades-STA}} & \multicolumn{4}{c}{\textbf{ActivityNet}} & \multicolumn{4}{c}{\textbf{YouCookII}}\\
        \cmidrule{3-14}
        & & R@0.3 & R@0.5 & R@0.7 & mIoU & R@0.3 & R@0.5 & R@0.7 & mIoU & R@0.3 & R@0.5 & R@0.7 & mIoU \\
        \midrule
        ExCL$^\dagger$ (orig.) & 6.9M & 65.10 & 44.10 & 22.60 & - & - & - & - & - & 44.20 & 28.00 & 14.60 & - \\
        ExCL (ours)  & 6.9M & 62.28 & \und{39.74} & 22.53 & 42.28 & 55.49 & 39.33 & 23.04 & 40.32 & 26.58 & 15.72 & 8.19 & 18.99 \\ 
        + BERT \Snowflake   & 6.9M & \und{62.93} & 38.44 & 22.23 & \und{42.38} & 57.21 & 39.66 & 23.79 & 41.45 & 26.63 & 16.15 & 8.51 & 18.87\\
        \, + Adapter      & 7.2M-16.8M & 61.59 & 37.15 & 21.51 & 41.06 & 59.35 & 41.27 & 24.86 & \und{42.83} & \und{28.47} & \und{17.75} & \und{9.02} & 19.89\\ 
        \, + Fine-tuning  & 116M & 61.75 & 38.36 & \und{23.44} & 42.00 & \und{59.10} & \und{41.83} & \und{25.42} & 42.36 & 28.18 & 16.84 & \und{9.02} & \und{20.08}\\
        \midrule
        TMLGA (orig.)     & 4.7M & 67.53 & 52.02 & 33.74 & 48.22 & 51.28 & 33.04 & 19.26 & 37.78 & 33.48 & 20.65 & 10.94 & 23.07 \\
        TMLGA (ours)      & 4.7M & 69.49 & 49.97 & 32.72 & 48.29 & 50.84 & 31.13 & 17.86 & 36.90 & 34.42 & 21.99 & 10.94 & 23.63 \\
        + BERT \Snowflake   & 4.7M & 70.08 & 49.92 & 31.42 & 48.34 & 52.10 & 32.57 & 18.64 & 37.63 & 34.77 & \und{23.05} & \und{12.49} & 24.42 \\
        \, + Adapter      & 5.6M - 14.6M & \und{71.40} & \und{52.53} & \und{33.82} & 49.57 & \und{53.98} & \und{35.20} & \und{20.43} & \und{38.88} & \und{36.08} & 22.77 & \und{12.49} & \und{25.19} \\ 
        \, + Fine-tuning  & 114M & 71.02 & \und{52.53} & 33.52 & \und{49.80} & 53.59 & 34.05 & 19.51 & 37.92 & 35.34 & 21.85 & 11.63 & 24.82 \\
        \midrule
        DORi (orig.)      & 10.4M & \bf\und{72.72} & \bf\und{59.65} & 40.56 & 53.28 & 57.89 & 41.49 & 26.41 & 42.78 & 43.36 & 30.47 & 18.24 & 30.46 \\
        DORi (ours.)      & 10.4M & 72.26 & 57.18 & 40.62 & 53.01 & 57.38 & 40.00 & 24.84 & 41.97 & 43.33 & 29.15 & 17.61 & 30.17 \\
        + BERT \Snowflake   & 10.4M & 71.83 & 57.15 & 39.22 & 52.49 & 58.86 & 40.86 & 25.50 & 42.97 & 42.27 & 29.90 & 18.38 & 29.92\\
        \, + Adapter      & 11.6M - 20.3M & 72.50  & 58.63 & \bf\und{40.97} & \bf\und{53.29} & \bf\und{60.81} & \bf\und{43.49} & \bf\und{27.86} & \bf\und{44.55} & \bf\und{46.79} & \bf\und{32.56} & \bf\und{19.87} & \bf\und{32.48} \\ 
        \bottomrule
    \end{tabular}
    }
    \caption{Overview of our results combining BERT and adapters with our selected prior work. Underlined results indicate the best performance within the method and dataset combination, while results in bold indicate the best performance within the dataset. }
    \label{table:bigtable}
    \vspace{-4mm}
\end{table*}

\paragraph{\RQ{1}: Effect of adding BERT to existing models}
To investigate this matter, we replace the non-contextualized word embeddings in our selected TVG models with BERT, which can be regarded as the current most widely-studied PLM \cite{rogersPrimerBERTologyWhat2020a,yangBERTRepresentationsVideo2020,chenUNITERUNiversalImageTExt2020,liOscarObjectSemanticsAligned2020}. 
We compare the original model performance with the performance when fine-tuning the PLM along with the TVG model training (fine-tuning), and when freezing the PLM (\Snowflake), training only the parameters of the TVG model. To ensure our implementations were correct, we also tested the original models (ours), achieving performance close to the reported in their respective papers.  

\begin{figure}[t]
\centering
\begin{subfigure}{0.95\linewidth}
  \centering
  \includegraphics[width=\linewidth]{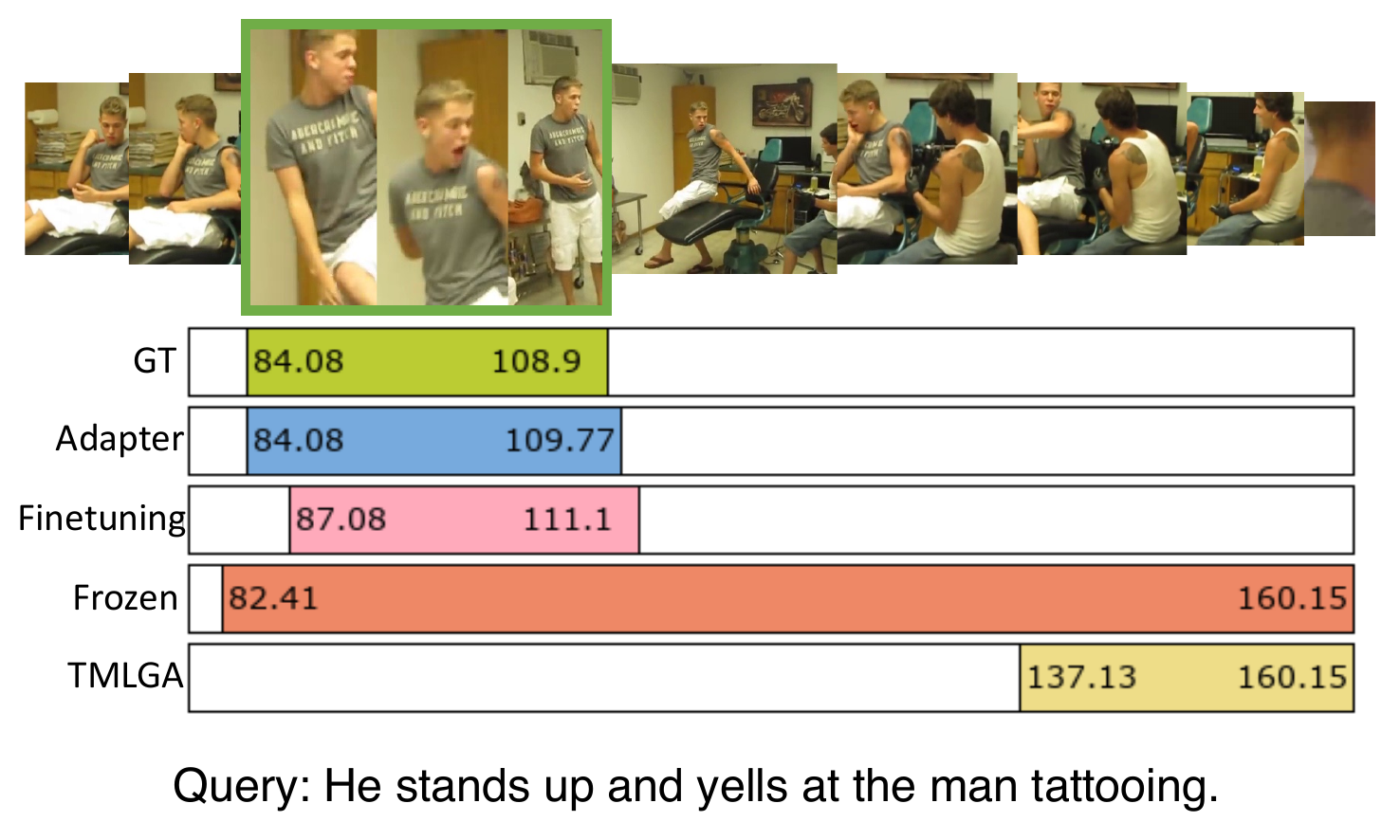}
  \label{fig:sub1}
\end{subfigure}%
\\
\begin{subfigure}{0.95\linewidth}
  \centering
  \includegraphics[width=\linewidth]{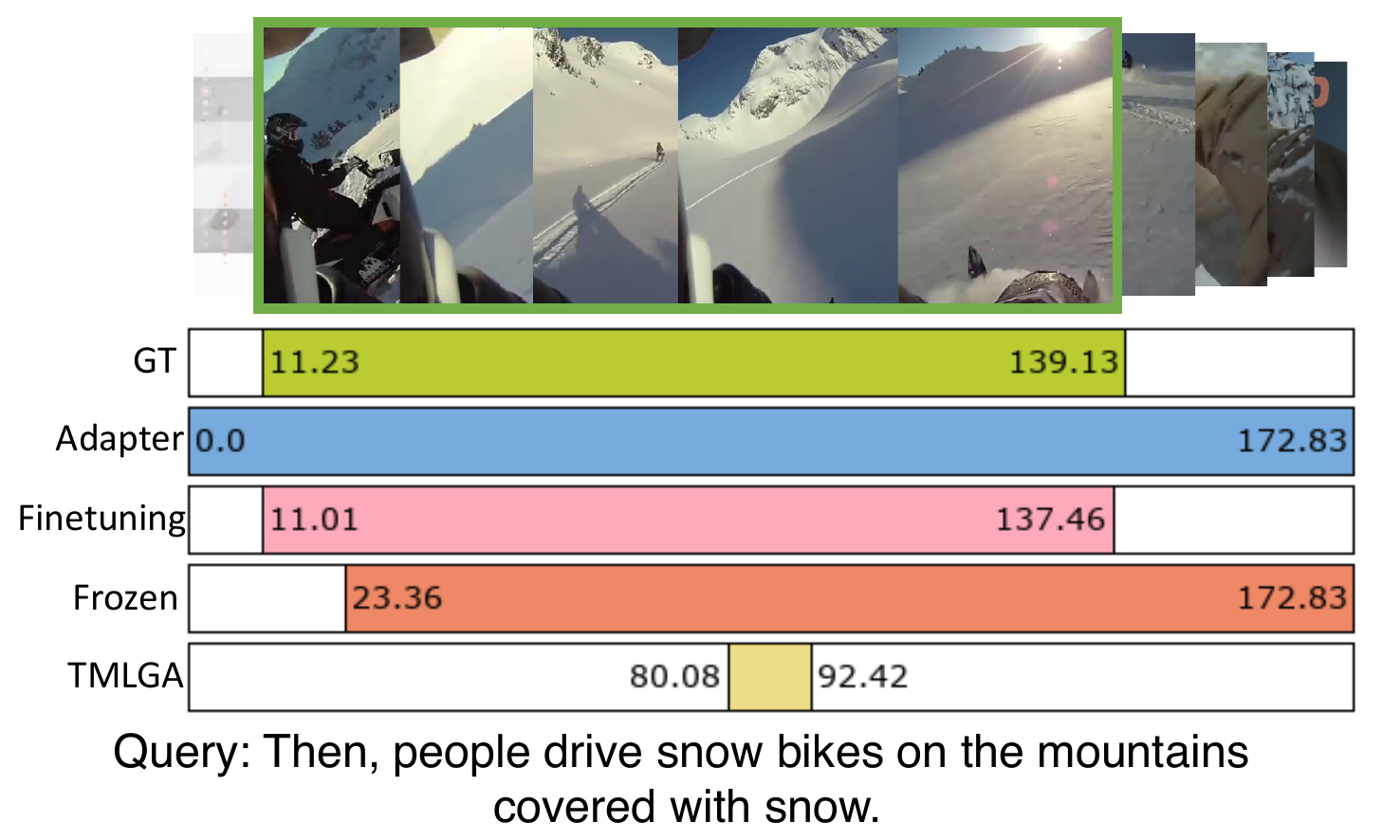}
  \label{fig:sub2}
\end{subfigure}
\vspace{-5mm}
\caption{Examples of success (top) and failure (bottom) of TMLGA with BERT on ActivityNet.}
\label{fig:examples}
\vspace{-5mm}
\end{figure}

Table~\ref{table:bigtable} shows the results of this combination. 
When introducing BERT to the models, we can see that full model fine-tuning leads to an average improvement of $1.38\%$ and $1.24\%$ in mIoU for ExCL and TMLGA, respectively. This result shows that the chosen TVG models can benefit from using PLMs. However, BERT adds over 100M parameters to be tuned. Such an increase is troublesome as we ran out of memory when trying to fine-tune BERT along with DORi.

One alternative to save on this computational cost is to freeze the PLM when training the TVG model. We can see that this strategy leads to an overall improvement in mIoU of $0.40\%$, $0.52\%$, and $0.08\%$ for ExCL, TMLGA, and DORi, respectively, when compared to the original model performance. This improvement is substantially smaller than when fine-tuning the PLM, with cases where using the frozen BERT leads to worse results than when using GloVe, such as with ExCL and DORi on YouCookII. Furthermore, we can see that the models benefited the least from the frozen PLM when tested on the Charades-STA dataset. This result could be due to queries in Charades-STA being less complex, so the word embeddings could already be enough to perform well. Nevertheless, the overall results indicate that fine-tuning is essential to getting the full potential of the PLM in our task.

\paragraph{\RQ{2}: Adapters as an alternative to PLM fine-tuning} 

To the best of our knowledge, there is no evidence to suggest whether adapters could bring benefits to the TVG task similar to what has been shown in other NLP tasks. Therefore, we seek to investigate if using adapters can be an effective alternative to full fine-tuning of the PLM models in TVG. We tested the adapters mentioned in Section~\ref{sec:adapters} for all three TVG models with BERT. 

Our best results are shown in Table~\ref{table:bigtable} (Adapter). For ExCL, the best adapters were \Pfeiffer{}, \Pfeiffer{}, and \Lora{}, for Charades-STA, ActivityNet and YouCookII, respectively; For TMLGA, the best were \Prefix{}, \Pfeiffer{}, and \Houlsby{}; and for DORi, \Inverse{}, \Prefix{}, and \Houlsby{}. Further results can be found in Appendix~\ref{sec:app:detailedBERT}. We also provide the visualization of success and failure examples in Figure~\ref{fig:examples} for the combination of TMLGA with BERT on ActivityNet, and refer the readers to the Appendix~\ref{sec:app:visualizations} for more visualizations.

Our results show that using adapters led to an overall improvement in mIoU of $0.70\%$, $1.71\%$ and $1.72\%$ for ExCL, TMLGA, and DORi, respectively, over the models' original performance. While the improvement from the adapters for ExCL is smaller than when doing full fine-tuning, adapters led to better performance with TMLGA. More importantly, adapters allowed a significant performance improvement for DORi, as training with them requires updating only 16\% of the parameters required for full fine-tuning. Furthermore, we can see that in some cases, using adapters leads to better performance than full fine-tuning, such as with ExCL and TMLGA on ActivityNet Captions.

In summary, these results indicate that despite not being tailored for the task of TVG, the adapters covered in this work can be an efficient alternative to the full fine-tuning of PLM models.

\begin{figure*}[t]
    \centering
    \includegraphics[width=\linewidth]{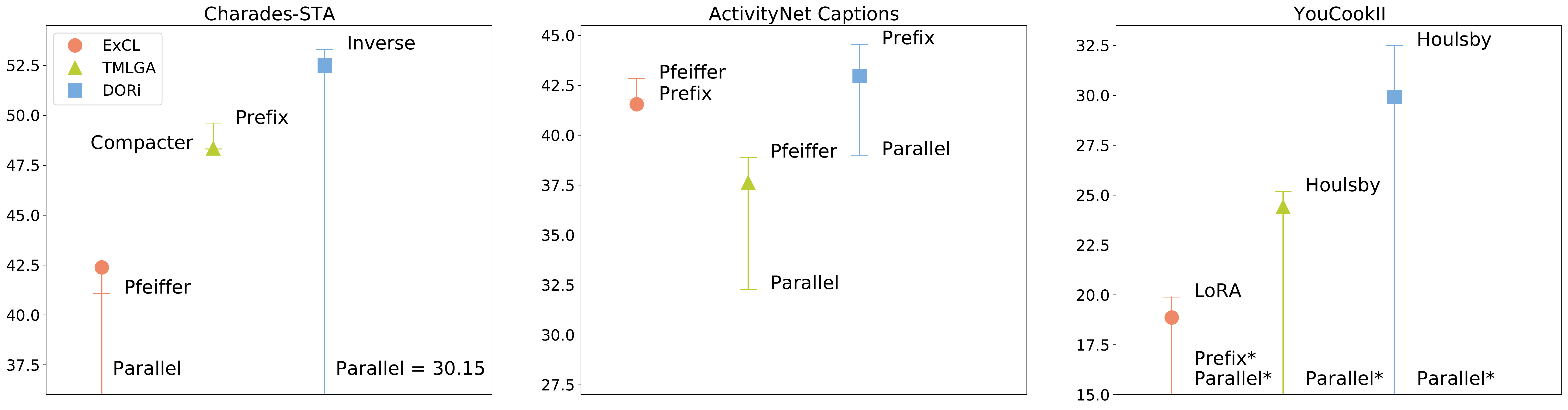}
    \caption{Best and worst performing adapters with BERT. The y-axis represents the performance in terms of mIoU. The circles, triangles and squares indicate the performance with frozen BERT, while the horizontal ticks show the performance of the models with the corresponding adapter. The * indicates cases where the models could not learn properly with the adapter.}
    \label{fig:bestworst}
    \vspace{-4mm}
\end{figure*}

\paragraph{\RQ{3}: Choice of adapter for TVG}
We were then naturally interested in studying whether there is a specific type of adapter that works best to solve our task. Therefore, we ranked the performance of each adapter for each dataset and model, focusing on cases that deliver the best and the worst performance.

The results of our analysis are summarized in Figure~\ref{fig:bestworst}. We can see that no single adapter can consistently offer the best performance. However, it is possible to see that the bottleneck adapters, such as \Houlsby{}, \Pfeiffer{}, and \Inverse{}, can deliver an overall better performance.

When controlling our results for each dataset, it is possible to see that the \Prefix{} and the \Inverse{} adapters worked well on Charades-STA, while the \Pfeiffer{} and \Houlsby{} adapters worked best for ActivityNet and YouCookII, respectively. We surmise this indicates that the adapter choice is likely more related to the training data than to the choice of the model itself. Controlling by model also shows interesting, distinctive patterns. For example, while the \Prefix{} adapter worked well for TMLGA on Charades-STA and DORi on ActivityNet, it performed poorly with ExCL on ActivityNet and YouCookII.

Another general pattern we could identify is that both the \Compacter{} and the \Parallel{} adapters did not work well overall, with the latter struggling to deliver good performance in all cases.
We think the number of additional parameters may play a role in this matter, as \Compacter{} is the adapter with the smallest number of added parameters (0.06M)
and it might be that too small to learn useful information for TVG.

Furthermore, the hyper-parameter search with the \Prefix{} and the \Parallel{} adapters was substantially more challenging. In particular, with the \Parallel{} adapter, we faced gradient explosions in many instances, as with DORi on Charades-STA, or the model converged to poor performance, as with all three models on YouCookII.

Finally, we believe another factor is the location where each adapter is inserted in the PLM architecture. All of the methods tested adapt to specific parts of the transformer layer
, except for the \Parallel~adapter, which adapts the whole transformer layer. Its consistent poor performance might indicate that adapting specific parts of the transformer layer is more beneficial for our task. 


\begin{table}[t]
    \centering
    \scalebox{0.75}{
    \begin{tabular}{l@{\hspace{0.15cm}} c@{\hspace{0.15cm}} c@{\hspace{0.15cm}} c@{\hspace{0.15cm}} c@{\hspace{0.15cm}} c@{\hspace{0.15cm}} c@{\hspace{0.15cm}}}
        \toprule
        \multirow{2}{*}{\bf Model} & \multicolumn{3}{c}{\bf Charades-STA} & \multicolumn{3}{c}{\bf ActivityNet} \\ 
        \cmidrule{2-7}
        & R@0.5 & R@0.7 & mIoU & R@0.5 & R@0.7 & mIoU\\
        \midrule
        TMLGA (ours)        & 49.97 & 32.72 & 48.29 & 31.13 & 17.86 & 36.90 \\
        \midrule
        + BERT \Snowflake    & 49.92 & 31.42 & 48.34 & 32.57 & 18.64 & 37.63 \\ 
        \, + \Pfeiffer      & 51.59 & 33.41 & 49.50 & \und{35.20} & \und{20.43} & \und{38.88}  \\
        \, + \Inverse       & \und{52.77} & \und{34.49} & 49.33 & 33.76 & 19.93 & 37.93 \\
        \, + Fine-tuning    & 52.53 & 33.52 & \und{49.80} & 34.05 & 19.51 & 37.92 \\
        \midrule
        + RoBERTa \Snowflake  & 51.34 & 33.49 & 48.91 & 33.80  & 19.62 & 37.89 \\
        \, + \Pfeiffer      & \textbf{\und{53.84}} & \textbf{\und{34.78}} & \textbf{\und{49.91}} & \und{35.27} & 20.26 & 38.77 \\
        \, + \Inverse       & 52.69 & 33.98 & 49.50 & 34.85 & \underline{20.46} & \underline{39.35} \\
        \, + Fine-tuning    & 53.15 & 33.33 & 49.77 & 33.21 & 19.69 & 38.47	\\
        \midrule
        + DeBERTa \Snowflake & 52.53 & 33.49 & 49.32 & 33.94 & 20.22 & 38.72 \\
        \, + \Pfeiffer      & \und{53.49} & \und{34.65} & \und{49.78} & 34.70 & 20.49 & 39.30\\
        \, + \Inverse       & 52.58 & 33.95 & 49.66 & \textbf{\und{35.45}} & \textbf{\und{20.66}} & \textbf{\und{39.71} }\\
        \, + Fine-tuning    & 53.44 & 33.44 & 49.63 & 33.78 & 20.12 & 38.92	\\
        \bottomrule
    \end{tabular}
    }
    \caption{Detailed results combining the TMLGA model with our three PLMs and adapters, tested on Charades-STA and ActivityNet Captions. Underlined results indicate the best performance within the model and dataset combination, while the results in bold indicate the best performance within the dataset.}
    \label{table:TMLGA-ANet-Charades}
    \vspace{-4mm}
\end{table}

\paragraph{\RQ{4}: Impact of different PLMs}
After the release of BERT, several pre-trained language encoder variations have been proposed. When tested on extensive NLP benchmarks, some of these models have proven to be able to consistently improve performance. In this context, we are interested in studying if such performance improvements also translate to better performance in our task. 

Thus, we study the performance of BERT, RoBERTa, and DeBERTa on two datasets, Charades-STA and ActivityNet, using TMLGA as a pivot. Our choice of TVG model is guided by our experiments with BERT, where TMLGA offered a good compromise in terms of performance improvements versus computational cost. Furthermore, our dataset selection is motivated by Charades-STA and ActivityNet having similar video contents but with different complexity queries, i.e., queries in Charades-STA are much simpler than in ActivityNet. We can verify this difference by observing the vocabulary size (748 vs. 9,744 tokens) and length of the queries (7.2 vs. 13.48 tokens per query). 

The results are summarized in Table~\ref{table:TMLGA-ANet-Charades}. For better readability, we only included two $\alpha$ bands. Further results can be seen in Appendix~\ref{sec:app:detailed_rob_deb}. We can first see that for all three PLMs, using adapters led to an overall performance improvement compared to the frozen PLM. Moreover, we noticed that \Pfeiffer~and the \Inverse~adapters seem to provide the best results in both datasets and all PLMs. 

Moreover, while DeBERTa performed the best for ActivityNet, as expected from its performance in NLP downstream tasks, its best performance was similar to BERT for Charades-STA.
We believe this result could be due to the simplicity of the queries in Charades-STA, which might not require all the additional information DeBERTa encodes. In addition, these results were obtained using the first version of DeBERTa, which uses the same type of tokenizer as BERT. We also tested the newer version of DeBERTa, which uses a sentencepiece-based tokenizer and incorporates other model improvements. However, this model performed much worse than the first version on our task showing that using sophisticated tokenizers does not necessarily improve results with simple sentence queries. 

On the other hand, ActivityNet has longer and more complex queries and our results indicate that in such cases, our task might benefit from using better PLMs. Nevertheless, the best results were achieved when using adapters.

\begin{table}[t]
    \centering
    \scalebox{0.75}{
        \begin{tabular}{l@{\hspace{0.08cm}} c@{\hspace{0.15cm}}c@{\hspace{0.15cm}} c@{\hspace{0.15cm}}c@{\hspace{0.15cm}}c@{\hspace{0.15cm}}c@{\hspace{0.15cm}} }
            \toprule
            \multirow{2}{*}{\textbf{Model}} & \multicolumn{3}{c}{\textbf{Charades-STA}} & \multicolumn{3}{c}{\textbf{ActivityNet}}\\ 
            \cmidrule{2-7}
            & R@0.5 & R@0.7 & mIoU & R@0.5 & R@0.7 & mIoU \\
            \midrule
            \midrule
            \multicolumn{7}{c}{Proposal-free} \\
            \midrule
            DORi (ours) &  57.18 & 40.62 & 53.01 & 40.00 & 24.84 & 41.97\\
             + DeBERTa \Snowflake &  58.17 & 40.94 & 52.73 & 41.65 & 25.82 & \und{43.64}  \\
             \, + Adapter      &  58.39 & \und{41.61} & \bf53.34 & \bf45.63 & \und{28.74} & \bf45.70 \\ 
            \midrule
            VSLNet & 54.19 & 35.22 & 50.02 & 43.22 & 26.16 & 43.19 \\ 
            CPNet &  \und{60.27} & 38.74 & 52.00 & 40.56 & 21.63 & 40.65\\ 
            CPN & 59.77 & 36.67 & \und{53.14} & \und{45.10} & 28.10 & \bf45.70 \\ 
            BCPN  & \bf61.77 & \bf43.91 & - &  44.53 & \bf30.11 & - \\
            \midrule
            \midrule
            \multicolumn{7}{c}{Proposal-based} \\
            \midrule
            MS-2D-TAN & \bf60.08 & \bf37.39 & - &  45.50 & 28.28 & - \\ 
            MSAT  &  - & - & - & 48.02 & 31.78 & - \\
            CPL  &  \und{49.24} & 22.39 & - &\bf55.73 & \bf31.37 & - \\
            MNM  &  47.31 & \und{27.28} & - & \und{48.59} & \und{29.26} & - \\

            \bottomrule
        \end{tabular}
    }
    \caption{Comparison between the best performing TVG model with DeBERTa and current state-of-the-art methods in TVG. Results for all compared methods were taken from their respective papers. The best results for each combination of method type (i.e., proposal-free or proposal-based) and dataset are indicated in bold, while the second-best results are underlined.} 
    \label{table:SOTA}
    \vspace{-4mm}
\end{table}

\paragraph{\RQ{5}: Comparison against state-of-the-art}
We finally compare our best-performing models against a selection of approaches from previous work. We achieved our best performance for the Charades-STA dataset by using DORi with DeBERTa+\Pfeiffer{}; and for the ActivityNet dataset, DORi with DeBERTa+\Inverse. Since the models used in this study are proposal-free, we mainly compare to proposal-free methods, i.e.,  VSLNet, CPN, CPNet and BCPN. Nevertheless, we are also interested in observing how our modifications perform against proposal-based methods, such as MS-2D-TAN~\cite{zhangMultiScale2DTemporal2021}, MSAT,
CPL, and MNM. 
We note that out of the proposal-free methods, only BCPN uses a PLM (BERT). As for the proposal-based methods, only MS-2D-TAN does not use any form of Transformers in its architecture, while MNM is the only one to use a fine-tuned PLM (DistilBERT). Finally, while MSAT and CPL use Transformers in their architecture to encode visual and textual information, they do not use PLMs.

The performance summary is shown in Table~\ref{table:SOTA}. For this analysis, we only consider two $\alpha$ bands for the thresholds as most proposal-based models do not report results at the $\alpha$ $0.3$.

First, we can see that the best-performing methods are proposal-free, with BCPN achieving the best performance in most of the considered metrics. Furthermore, looking at the mean tIoU, we can see that DORi already performs well against the other methods. Replacing GloVe embeddings with DeBERTa and using adapters for training provides a significant performance boost, delivering results equivalent to state-of-the-art models.

Moreover, it is interesting to see that while proposal-based methods such as CPL and MNM performed well against all methods on ActivityNet, they could not outperform methods without Transformers on the Charades-STA dataset. In contrast, DORi with DeBERTa and adapters achieved a more balanced performance among both datasets, showing a clear advantage against these methods on the Charades-STA dataset.  

Therefore, our results show how TVG models can greatly benefit from adequately incorporating PLMs and making use of parameter-efficient techniques, performing well on datasets with different complexity levels in terms of queries, and achieving results comparable to state-of-the-art methods.

\section{Conclusions}
This paper studied the effects of PLMs in the TVG task and assessed the applicability of NLP parameter-efficient training alternatives based on adapters. We coupled BERT, RoBERTa, and DeBERTa, with a selection of previous TVG works, i.e., ExCL, TMLGA, and
DORi, and tested different adapters to reduce the impact of
the additional parameters. 
Our results showed that, by only changing the query representation using PLMs, TVG models can greatly benefit from such integration, especially when PLMs are fine-tuned, highlighting the importance of the query representation in this task.
Moreover, we verified that adapters are an effective alternative to full fine-tuning, even though they were not tailored for our task. They saved on computational cost, allowing improvements for larger TVG models, such as DORi, and also delivered results comparable to SOTA models. Finally, we observed that while \Parallel{} adapters struggled to learn in this task, bottleneck adapters such as \Houlsby{} and \Pfeiffer{} performed across all tested TVG models and datasets.

\section*{Limitations}
In this work, we studied the effects of large pre-trained models in the temporal video grounding task and investigated the applicability of NLP adapters for a parameter-efficient integration. While we believe our results show the efficacy of incorporating better language models in TVG models, it is important to note that we primarily focused on proposal-free TVG models and thus have no evidence to suggest such improvement would be observed in proposal-based models.

Furthermore, as our main goal was to investigate how the chosen models' performance varied when only changing the text encoding models, we compared state-of-the-art models using different visual features. While it would be interesting and insightful to check their performance when using the same features as our chosen models (i.e., I3D), such experiments are out of the scope of this study.

Moreover, although language adapters can be stacked before a task adapter for training on the task in a new language, we have only experimented with queries in English. It would be interesting to investigate if language adapters could be applied to TVG in different languages.

Finally, as for hardware requirements, our experiments were performed on a single 40-GB NVIDIA A100 GPU from a large cluster, and we spent about 400 USD on our experimental setup. While experiments with ExCL and TMLGA can be run on smaller GPUs with no significant increase in training time (i.e., we tested with a 16-GB NVIDIA V100 GPU), for DORi, due to the size of the input features and number of training parameters, we recommend using a GPU with at least 32GB of memory.

\section*{Ethics Statement}
This work does not present any direct ethical issues. Code for TMLGA and DORi were released by their respective authors. The datasets used to evaluate our proposed approach are open-access, and data characteristics relevant to our task were described in the experimental evaluation section. References for further information on each dataset were included in the paper.

\section*{Acknowledgements}

This paper is based on results obtained from the project JPNP20006, commissioned by the New Energy and Industrial Technology Development Organization (NEDO). For experiments, computational resource of AI Bridging Cloud Infrastructure (ABCI) provided by National Institute of Advanced Industrial Science and Technology (AIST) was used.

\bibliography{anthology,custom}
\bibliographystyle{acl_natbib}

\appendix

\clearpage

\section{Detailed Results with BERT}
\label{sec:app:detailedBERT}

We report the detailed results with BERT for ExCL, TMLGA, and DORi in Tables~\ref{table:excl_BERT}, ~\ref{table:tmlga_BERT} and ~\ref{table:dori_bert}, respectively. We also report the average runtime for training and inference of each TVG model when combined with BERT and the selected adapters in Table~\ref{table:time}. However, we note that these values also include the loading time of the samples, which varied accordind to the usage of the cluster during each experiment. 

\subsection{Hyper-parameters}

To reproduce the results of the original models, we started by using the hyper-parameters reported in the respective papers, achieving close to reported performance for ExCL and TMLGA. Reproducing DORi results was slightly more challenging, where we had to experiment with different weight decays, batch sizes and steps for the learning scheduler.

When adding the PLMs and the adapters, we first started with the same set of hyper-parameters of the original model. In general, it was necessary to slightly change them for the model to properly learn.
The hyper-parameters used to train ExCL are specified in Table~\ref{table:excl_BERT_Charades_param}
, Table~\ref{table:excl_BERT_Anet_params}, and Table~\ref{table:excl_BERT_Youcookii_params} for Charades-STA, ActivityNet Captions and YouCookII datasets, respectively; to train TMLGA, in Table~\ref{table:tmlga_BERT_Charades_param}, Table~\ref{table:tmlga_BERT_Anet_param}, and Table~\ref{table:tmlga_BERT_YouCookII_param}, respectively; and to train DORi, in Table~\ref{table:dori_BERT_Charades_param}, Table~\ref{table:dori_BERT_Anet_param}, and Table~\ref{table:dori_BERT_YouCookII_param}, respectively.

\subsection{Notes on training}

Specifically for the fine-tuning of BERT with TMLGA on the Charades-STA dataset, we report the results obtained by applying the linear warm-up to all parameters, including non-BERT ones, as this strategy led to the best results. Moreover, when training DORi with BERT and the \Parallel{} adapter on the Charades-STA dataset, we tested different weight decays, but the gradient exploded in all cases. The reported results were obtained with a weight decay of $1e-5$, the best result before the gradient exploded. Finally, we could not find a proper hyper-parameter combination so that the models could learn on the YouCookII with the \Parallel{} adapter. 

\section{Detailed Results with RoBERTa and DeBERTa}
\label{sec:app:detailed_rob_deb}

Detailed results with RoBERTa and DeBERTa with TMLGA on the Charades-STA and ActivityNet Captions datasets can be found on table~\ref{table:TMLGA-ANet-Charades_full}. This tables is an expansion of the results shown in table~\ref{table:TMLGA-ANet-Charades} in the main text.
We did not include results with \Prefix~adapters on DeBERTa due to an implementation error on the adapter-transformers library\footnote{Version 3.1.0a0 of the adapter-transformers library.}. Furthermore, we also note that in our experiments with the \Parallel~adapter, all the models saturated and reached the same performance on ActivityNet, without properly learning.

\subsection{Hyper-parameters}

We report the hyper-parameters used to train TMLGA with RoBERTa on Charades-STA and ActivityNet Captions on Table~\ref{table:tmlga_roberta_Charades_param} and Table~\ref{table:tmlga_roberta_Anet_param}, respectively. We also report the hyper-parameters used to train TMLGA with DeBERTa on both datasets on Table~\ref{table:tmlga_deberta_charades_param} and Table~\ref{table:tmlga_deberta_anet_param}.




\subsection{Hyper-parameters for best result}
Our best results on the Charades-STA dataset was achieved by training DORi with DeBerta, using the \Pfeiffer{} adapter, while the best results on the ActivityNet Captions was achieved by training DORi with DeBERTa, using the \Inverse{} adapter. The hyper-parameters used to achieve these results can be found in Table~\ref{table:dori_deberta}. 


\section{Qualitative Results}
\label{sec:app:visualizations}
\begin{figure}[t]
\centering
\begin{subfigure}{\linewidth}
  \centering
  \includegraphics[width=\linewidth]{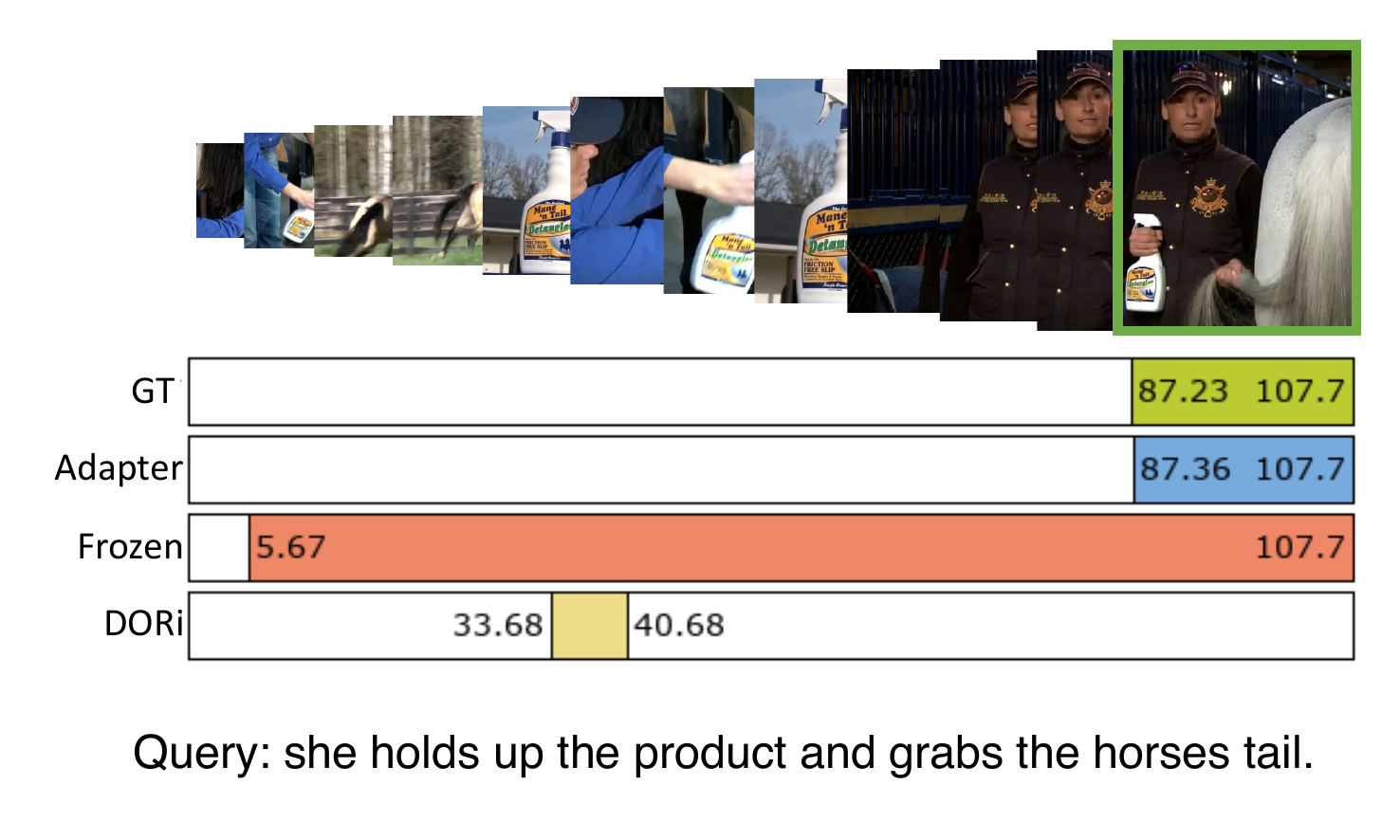}
  \label{fig:sub1_dori}
\end{subfigure}%
\\
\begin{subfigure}{\linewidth}
  \centering
  \includegraphics[width=\linewidth]{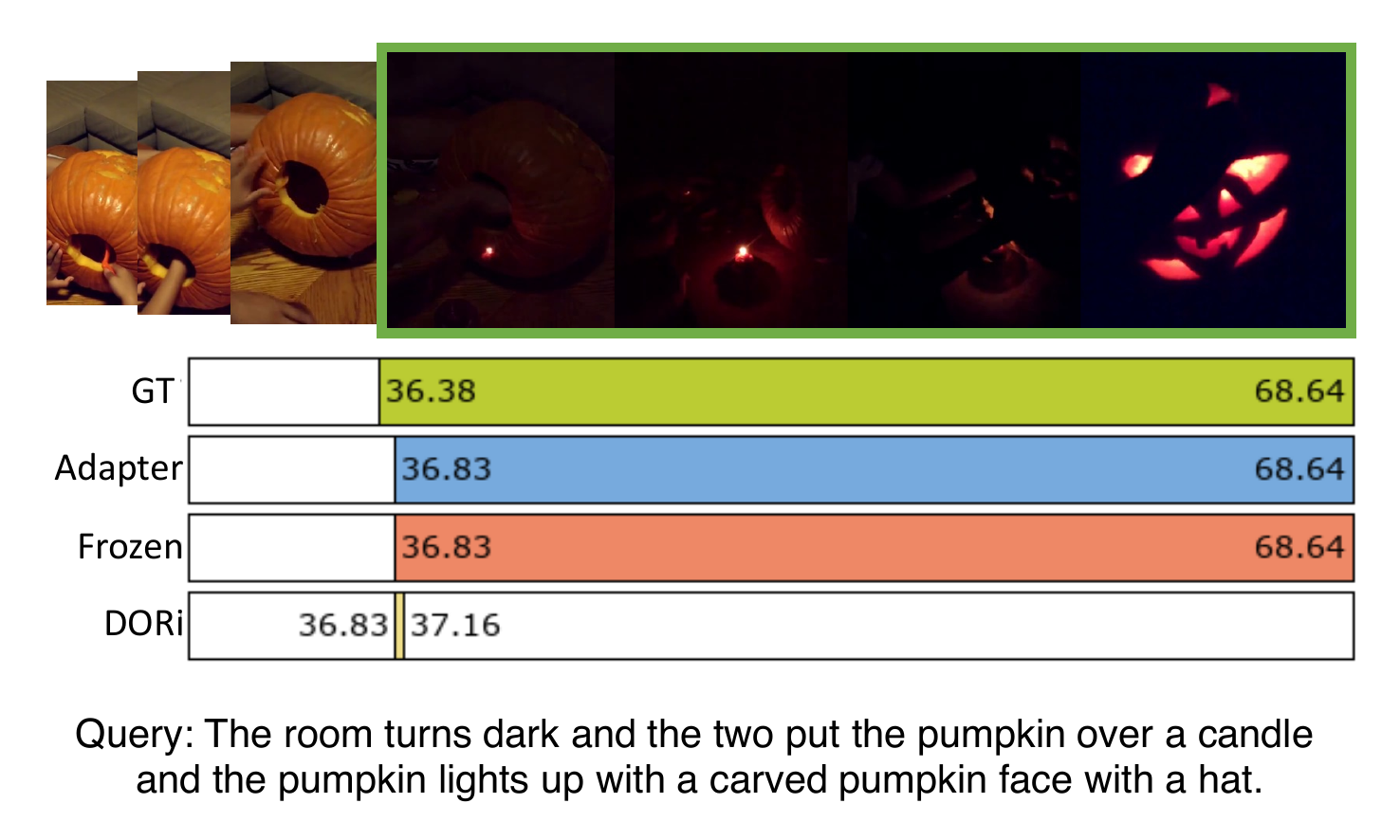}
  \label{fig:sub2_dori}
\end{subfigure}
\vspace{-5mm}
\caption{Examples showing the effects of BERT and adapters with DORi on the ActivityNet. The top image shows an example of a significant performance improvement only when using adapters. On the other hand, the bottom image shows an example where the frozen PLM was sufficient to correctly identify the video segment represented by the query.}
\label{fig:examples_DORI}
\vspace{-5mm}
\end{figure}

\begin{figure}[t]
\centering
\begin{subfigure}{\linewidth}
  \centering
  \includegraphics[width=\linewidth]{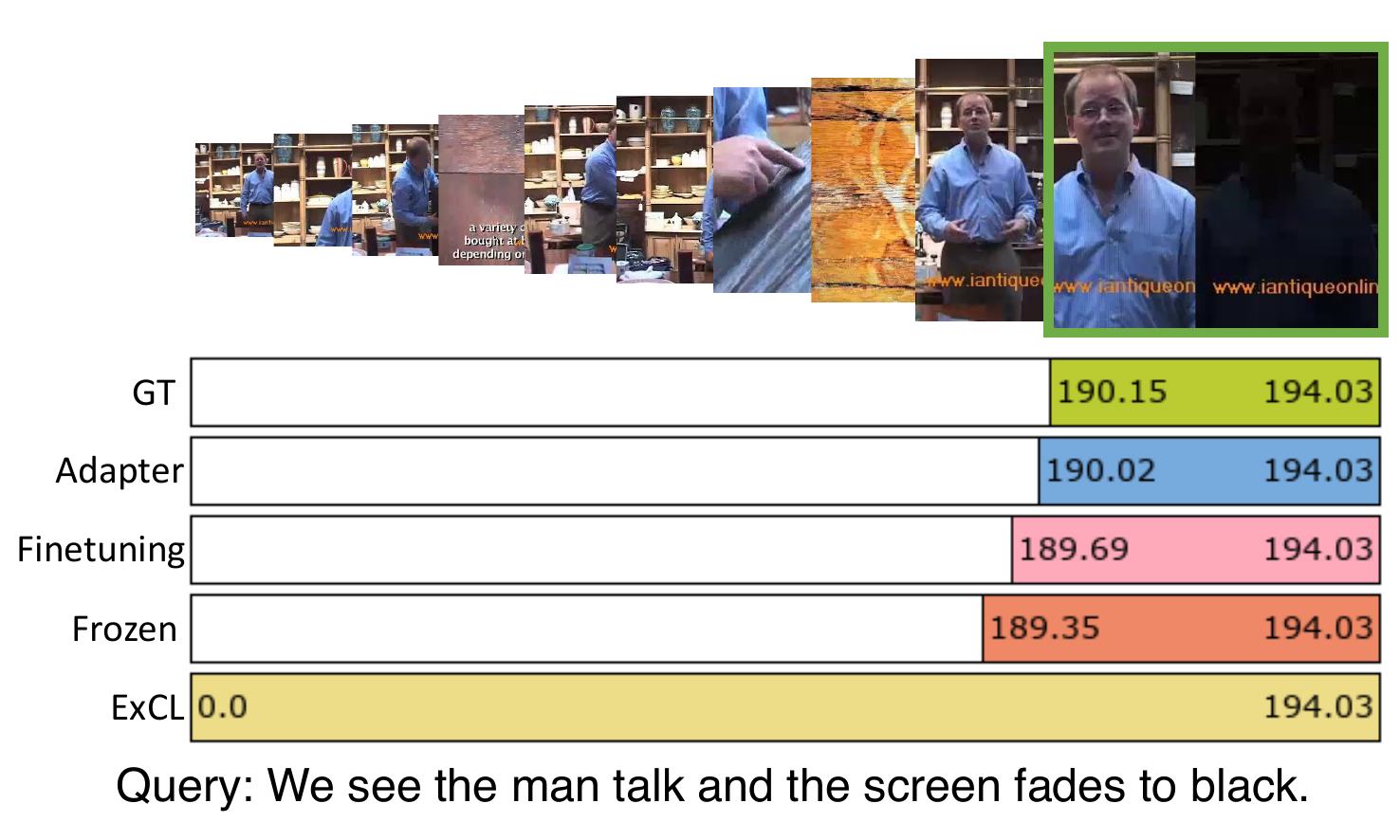}
  \label{fig:sub1_excl}
\end{subfigure}%
\\
\begin{subfigure}{\linewidth}
  \centering
  \includegraphics[width=\linewidth]{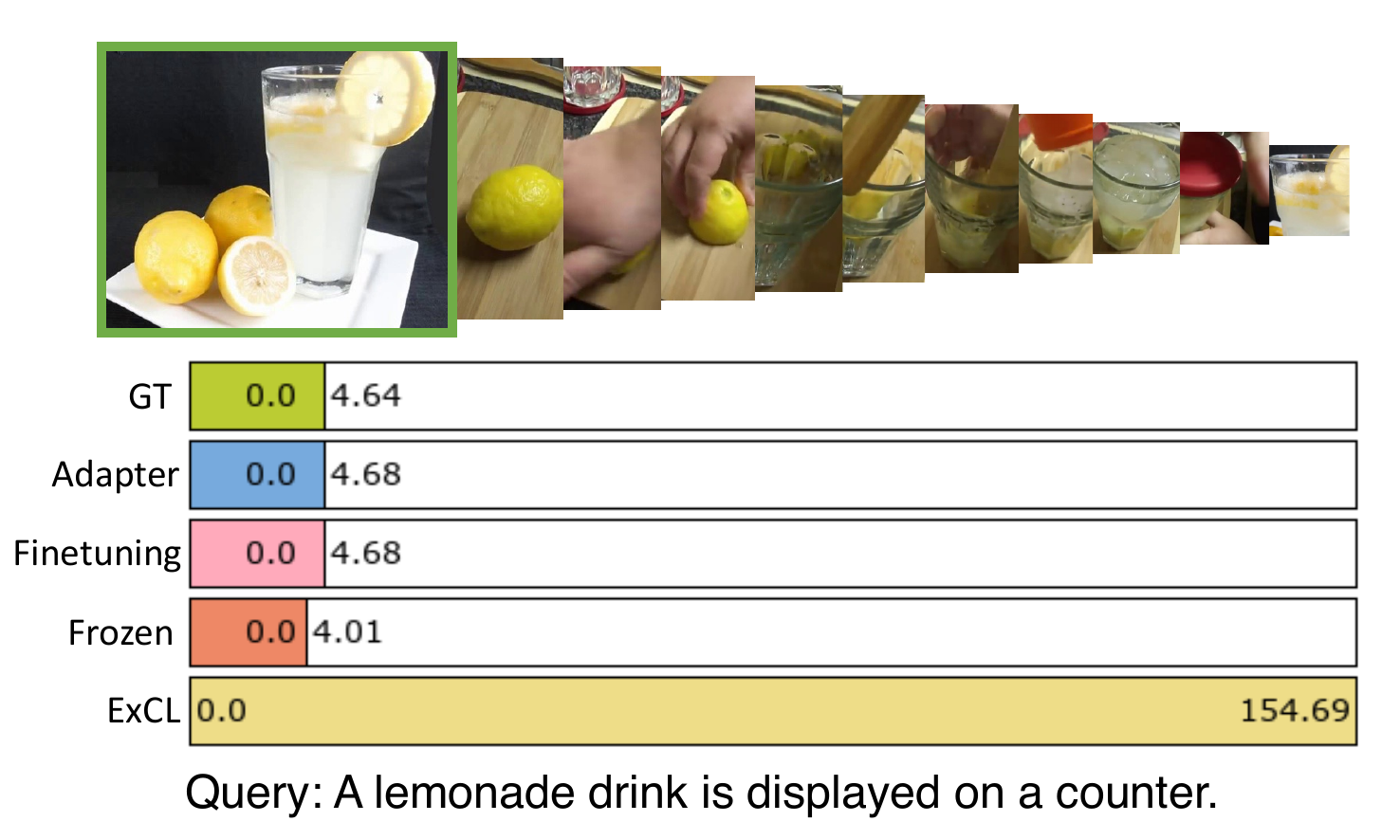}
  \label{fig:sub2_excl}
\end{subfigure}
\vspace{-5mm}
\caption{Examples showing the effects of BERT and adapters with ExCL on the ActivityNet. Both images show the significant impact of using PLMs on this model's performance, drastically improving results when incorporating frozen BERT. However, while in the top image, the best results were achieved by training using adapters, in the bottom image, we can see that the frozen PLM was sufficient to solve the respective query.}
\label{fig:examples_EXCL}
\vspace{-5mm}
\end{figure}

Finally, we provide a few examples of failure and success of each analyzed TVG model along with BERT. Figure~\ref{fig:examples_DORI} shows examples for DORi and Figure~\ref{fig:examples_EXCL} shows examples for ExCL.

\clearpage

\begin{table*}[t]

    \centering
    \scalebox{0.8}{
    
    \begin{tabular}{l c@{\hspace{0.15cm}} c@{\hspace{0.15cm}}c@{\hspace{0.15cm}}c@{\hspace{0.15cm}}c@{\hspace{0.15cm}} c@{\hspace{0.15cm}}c@{\hspace{0.15cm}}c@{\hspace{0.15cm}}c@{\hspace{0.15cm}} c@{\hspace{0.15cm}}c@{\hspace{0.15cm}}c@{\hspace{0.15cm}}c@{\hspace{0.15cm}}}
        \toprule
        \multirow{2}{*}{\textbf{Method}} & \multirow{2}{*}{\textbf{Params.}} & \multicolumn{4}{c}{\textbf{Charades-STA}} & \multicolumn{4}{c}{\textbf{ActivityNet}} & \multicolumn{4}{c}{\textbf{YouCookII}}\\
        \cmidrule{3-14}
        & & R@0.3 & R@0.5 & R@0.7 & mIoU & R@0.3 & R@0.5 & R@0.7 & mIoU & R@0.3 & R@0.5 & R@0.7 & mIoU \\
        \midrule
        ExCL (ours)             & \ \ 6.9M  & 62.28 & \bf39.74 & 22.53 & 42.28 & 55.49 & 39.33 & 23.04 & 40.32 & 26.58 & 15.72 & \ \ 8.19 & 18.99 \\ 
        + BERT \Snowflake         & \ \ 6.9M  & \bf62.93 & 38.44 & 22.23 & \bf42.38 & 57.21 & 39.66 & 23.79 & 41.45 & 26.63 & 16.15 & \ \ 8.51 & 18.87\\
        \, +  \Houlsby          & \ \ 8.7M  & 60.08 & 36.26 & 20.83 & 40.22 & 57.79 & 39.77 & 23.80 & 41.95 & 25.77 & 15.12 & \ \ 8.36 & 18.52 \\
        \, +  \Pfeiffer         & \ \ 7.8M  & \und{61.59} & 37.15 & \und{21.51} & \und{41.06} & \und{\bf59.35} & \und{41.27} & 24.86 & \und{\bf42.83} & 27.18 & 16.15 & \ \ 8.88  & 19.46 \\
        \, +  \Inverse          & \ \ 8.1M & 60.81 & \und{37.69} & 21.10 & 40.85 & 57.58 & 40.34 & 24.79 & 42.06 & 26.72 & 16.44 & \ \ 8.79  & 19.43 \\
        \, +  \Prefix           & 16.8M     & 60.54 & 35.24 & 20.22 & 40.98 & 57.49 & 39.55 & 24.10 & 41.77 & \ \ 4.93  & \ \ 1.83  &\ \ 0.66  & \ \ 0.06  \\
        \, +  \Compacter        & \ \ 7.0M & 59.19 & 35.38 & 20.97 & 39.99 & 57.82 & 39.31 & 23.86 & 41.89 & 27.81 & 16.67 & \ \  \und{\bf9.08}  & 19.57 \\
        \, +  \Lora             & \ \ 7.2M & 60.94 & 36.32 & 21.48 & 40.94 & 57.94 & 40.64 & \und{25.05} & 42.49 & \und{\bf28.47} & \und{\bf17.75} & \ \ 9.02  & \und{19.89} \\
        \, +  \Parallel         & 14.0M     & 48.52 & 16.61 & \ \ 9.44 & 33.77 & 58.60 & 40.99 & 24.73 & 42.26 & \ \ 6.41  & \ \ 2.58  & \ \ 1.12  & \ \ 0.06 \\
        \, + Fine-tuning        & 116M      & 61.75 & 38.36 & \bf23.44 & 42.00 & 59.10 & \bf41.83 & \bf25.42 & 42.36 & 28.18 & 16.84 & \ \ 9.02 & \bf20.08\\
        
        \bottomrule
    \end{tabular}
    }
    \caption{Detailed results for ExCL using BERT. Underlined results indicate the best adapter performance, while results in bold indicate the best performance within the dataset.}
    \label{table:excl_BERT}
    
\end{table*}

\begin{table*}[t]
    
    \centering
    \scalebox{0.8}{
    \begin{tabular}{l c@{\hspace{0.15cm}} c@{\hspace{0.15cm}}c@{\hspace{0.15cm}}c@{\hspace{0.15cm}}c@{\hspace{0.15cm}} c@{\hspace{0.15cm}}c@{\hspace{0.15cm}}c@{\hspace{0.15cm}}c@{\hspace{0.15cm}} c@{\hspace{0.15cm}}c@{\hspace{0.15cm}}c@{\hspace{0.15cm}}c@{\hspace{0.15cm}}}
        \toprule
        \multirow{2}{*}{\textbf{Method}} & \multirow{2}{*}{\textbf{Params.}} & \multicolumn{4}{c}{\textbf{Charades-STA}} & \multicolumn{4}{c}{\textbf{ActivityNet}} & \multicolumn{4}{c}{\textbf{YouCookII}}\\
        \cmidrule{3-14}
        & & R@0.3 & R@0.5 & R@0.7 & mIoU & R@0.3 & R@0.5 & R@0.7 & mIoU & R@0.3 & R@0.5 & R@0.7 & mIoU \\
        \midrule
        TMLGA (ours)      & \ \ 4.7M   & 69.49 & 49.97 & 32.72 & 48.29 & 50.84 & 31.13 & 17.86 & 36.90 & 34.42 & 21.99 & 10.94 & 23.63 \\
        + BERT \Snowflake   & \ \ 4.7M   & 70.08 & 49.92 & 31.42 & 48.34 & 52.10 & 32.57 & 18.64 & 37.63 & 34.77 & \bf23.05 & \und{\bf12.49} & 24.42 \\
        \, +  \Houlsby    & \ \ 8.7M   & 70.97 & 51.69 & 33.84 & 49.31 & \und{\bf53.98} & 34.13 & 19.48 & 38.57 & \und{\bf36.08} & \und{22.77} & \und{\bf12.49} & \und{\bf25.19} \\
        \, +  \Pfeiffer   & \ \ 7.8M   & \und{\bf71.64} & 51.59 & 33.41 & 49.50 & \und{\bf53.98} & \und{\bf35.20}  & \und{\bf20.43} & \und{\bf38.88} & 34.31 & 21.76 & 11.31 & 23.54 \\
        \, +  \Inverse    & \ \ 8.1M   & 71.02 & 52.77 & \und{\bf34.49} & 49.33 & 52.47 & 33.76 & 19.93 & 37.93 & 35.40 & 22.05 & 11.14 & 24.44 \\
        \, +  \Prefix     & 16.8M  & 71.40 & \und{\bf52.53} & 33.82 & \und{49.57} & 53.61 & 34.03 & 19.89 & 38.34 & 18.36 & 10.17 &\ \ 4.64  & 13.62 \\
        \, +  \Compacter  & \ \ 7.0M   & 70.27 & 49.73 & 31.37 & 48.31 & 52.15 & 33.85 & 19.62 & 37.78 & 35.34 & 22.25 & 11.05 & 24.13 \\
        \, +  \Lora       & \ \ 7.2M   & 70.94 & 50.24 & 32.15 & 48.81 & 51.90  & 32.81 & 18.77 & 37.37 & 35.88 & 22.65 & 11.91 & 24.51 \\
        \, +  \Parallel   &  14.0M  & 70.48 & 51.64 & 33.66 & 49.33 & 43.50  & 23.20  & 11.59 & 32.29 &\ \ 5.98  &\ \ 2.36  &\ \ 1.00   &\ \ 0.06 \\
        \, + Fine-tuning  & \ \  114M & 71.02 & \bf52.53 & 33.52 & \bf49.80 & 53.59 & 34.05 & 19.51 & 37.92 & 35.34 & 21.85 & 11.63 & 24.82 \\
        
        \bottomrule
    \end{tabular}
    }
    \caption{Detailed results for TMLGA using BERT. Underlined results indicate the best adapter performance, while results in bold indicate the best performance within the dataset.}
    \label{table:tmlga_BERT}
\end{table*}

\begin{table*}[t]
    \centering
    \scalebox{0.8}{
    \begin{tabular}{l c@{\hspace{0.15cm}} c@{\hspace{0.15cm}}c@{\hspace{0.15cm}}c@{\hspace{0.15cm}}c@{\hspace{0.15cm}} c@{\hspace{0.15cm}}c@{\hspace{0.15cm}}c@{\hspace{0.15cm}}c@{\hspace{0.15cm}} c@{\hspace{0.15cm}}c@{\hspace{0.15cm}}c@{\hspace{0.15cm}}c@{\hspace{0.15cm}}}
        \toprule
        \multirow{2}{*}{\textbf{Method}} & \multirow{2}{*}{\textbf{Params.}} & \multicolumn{4}{c}{\textbf{Charades-STA}} & \multicolumn{4}{c}{\textbf{ActivityNet}} & \multicolumn{4}{c}{\textbf{YouCookII}}\\
        \cmidrule{3-14}
        & & R@0.3 & R@0.5 & R@0.7 & mIoU & R@0.3 & R@0.5 & R@0.7 & mIoU & R@0.3 & R@0.5 & R@0.7 & mIoU \\
        \midrule
        DORi (ours.)      & 10.4M & 72.26 & 57.18 & 40.62 & 53.01 & 57.38 & 40.00 & 24.84 & 41.97 & 43.33 & 29.15 & 17.61 & 30.17 \\
        + BERT \Snowflake   & 10.4M & 71.83 & 57.15 & 39.22 & 52.49 & 58.86 & 40.86 & 25.50 & 42.97 & 42.27 & 29.90 & 18.38 & 29.92\\
        \, +  \Houlsby    & \ \ 8.7M   & 72.72 & 57.58 & 40.59 & 53.16 & 60.71 & 43.18 & 26.97 & 43.93 & \und{\bf46.79} & \und{\bf32.56} & \und{\bf19.87} & \und{\bf32.48} \\
        \, +  \Pfeiffer   & \ \ 7.8M   & 72.28 & 58.49 & 40.89 & 53.13 & 60.67 & \und{\bf43.53} & 27.33 & 44.30 & 44.96 & 31.90  & 19.39 & 31.48 \\
        \, +  \Inverse    & \ \ 8.1M   & 72.50 & \und{\bf58.63} & \und{\bf40.97} & \und{\bf53.29} & \und{\bf61.01} & 43.90 & 27.68 & 44.32 & 45.50 & 30.76 & 19.13 & 31.48 \\
        \, +  \Prefix     & 16.8M  & 71.99 & 57.69 & 40.67 & 52.94 & 60.81 & 43.49 & \und{\bf27.86} & \und{\bf44.55} & 45.59 & 31.90 & 19.44 & 31.53\\
        \, +  \Compacter  & \ \ 7.0M   & \und{\bf72.63} & 57.98 & 40.83 & 52.93 & 60.73 & 43.03 & 27.40  & 44.31 & 43.81 & 30.13 & 18.36 & 30.58 \\
        \, +  \Lora       & \ \ 7.2M   & 70.73 & 57.31 & 39.76 & 51.89 & 60.90 & 43.34 & 27.46 & 44.42 & 45.42 & 31.44 & 19.47 & 31.56 \\
        \, +  \Parallel   & 14.0M  & 42.18 &\ \ 9.65 & \ \ 5.08 & 30.15 & 52.94 & 35.23 & 21.92 & 39.00 & \ \ 5.01  &\ \ 1.75  & \ \ 0.66  & \ \ 0.06  \\
        \, + Fine-tuning  & 120M & \multicolumn{4}{c}{OOM} & \multicolumn{4}{c}{OOM} & \multicolumn{4}{c}{OOM} \\
        
        \bottomrule
    \end{tabular}
    }
    \caption{Detailed results for DORi using BERT. Underlined results indicate the best adapter performance, while results in bold indicate the best performance within the dataset.}
    \label{table:dori_bert}
\end{table*}

\begin{table*}[t]
\centering
    \scalebox{0.8}{
    \begin{tabular}{l@{\hspace{0.15cm}} c@{\hspace{0.15cm}} c@{\hspace{0.15cm}} c@{\hspace{0.15cm}} c@{\hspace{0.15cm}} c@{\hspace{0.15cm}} c@{\hspace{0.15cm}}}
        \toprule
        \multirow{2}{*}{\textbf{Method}} & \multicolumn{2}{c}{\textbf{Charades-STA}} & \multicolumn{2}{c}{\textbf{ActivityNet}} & \multicolumn{2}{c}{\textbf{YouCookII}}\\
        \cmidrule{2-7}
        & Train & Inference & Train & Inference & Train & Inference\\
        \midrule
        ExCL (ours)  & 33.40 $\pm$ 0.88 &	10.70	$\pm$ 0.97 &\ \ 814.00 $\pm$	79.44 &	425.00 $\pm$	551.66 &	170.37	$\pm$ \ \ 
 0.92 &	63.85	$\pm$ 61.79\\ 
        + BERT \Snowflake   & 23.91 $\pm$ 1.97 & \ \ 7.83 $\pm$ 0.38 &\ \ 829.40 $\pm$ 39.20 & 151.80 $\pm$ \ \ \ 7.39 & 169.53 $\pm$ \ \ 
 1.01 & 39.44 $\pm$ \ \ 0.62\\
        \, + Adapter      & 32.18 $\pm$ 1.32 & 10.00 $\pm$ 0.47 &\ \ 833.00 $\pm$ 25.31 & 102.85 $\pm$ \ \ 11.52 & \ \  94.33 $\pm$ \ \ 
 0.88 & 21.81 $\pm$\ \  0.60\\ 
        \, + Fine-tuning  & 45.40 $\pm$ 2.67 & 10.10 $\pm$ 0.78 &   1883.00 $\pm$ 72.12 & 180.00 $\pm$ \ \ \ 1.41 & 306.18 $\pm$ 66.19 & 36.27 $\pm$ \ \ 1.73\\
        \midrule
        TMLGA (ours)      &  25.04 $\pm$ 0.74 & 15.44 $\pm$ 0.82 &  884.62 $\pm$ \ \ 10.56 & 123.75 $\pm$ \ \ 7.36 & 140.90 $\pm$ \ \  2.02 & 43.20 $\pm$ 1.470\\
        + BERT \Snowflake   &  25.50 $\pm$ 0.85 & 15.93 $\pm$ 0.73 & 454.18 $\pm$ \ \ \ 2.40 & \ \ 82.00 $\pm$ \ \ 1.00 & 163.27 $\pm$ \ \ 
 9.66 & 50.72 $\pm$ 47.19\\
        \, + Adapter      &  29.65 $\pm$ 0.77 & 17.95 $\pm$ 0.78 &  658.19 $\pm$ \ \ 24.41 & 136.44 $\pm$ \ \ 4.24 & 166.58 $\pm$ 10.26 & 49.33 $\pm$ 47.13\\ 
        \, + Fine-tuning  & 34.89 $\pm$ 0.93 & 16.55 $\pm$ 0.72 &   768.00 $\pm$ 219.64 & 137.92 $\pm$ 28.16 & 172.46 $\pm$ \ \  1.51 & 37.61 $\pm$ \ \ 0.87\\
        \midrule
        DORi (ours.)      & 631.24 $\pm$ 1.13 & 138.20 $\pm$ 0.75 & 6704.00 $\pm$ \ \  \ \ 1.41 & 1040.00 $\pm$ \ \ 4.24 & 2101.00 $\pm$ 81.99 & 845.88 $\pm$ 1054.53 \\
        + BERT \Snowflake   & 607.83 $\pm$ 1.52 & 134.67 $\pm$ 0.72 & 6599.00 $\pm$ \ \  21.66 & 1357.33 $\pm$ 23.46 & 2221.50 $\pm$ 72.83 & 667.50 $\pm$ 1593.11\\
        \, + Adapter      &  633.14 $\pm$ 1.34 & 130.14 $\pm$ 0.69 & 7415.00 $\pm$ 123.30 & 1405.75 $\pm$ 34.35 & 2279.50 $\pm$ 15.93 & 403.25 $\pm$ \ \ \ 29.06\\ 

        \bottomrule
    \end{tabular}
    }
    \caption{Average runtime in seconds for training and evaluating each selected TVG model when combining with BERT and adapters.}
    \label{table:time}
    \vspace{-4mm}
\end{table*}



    




\begin{table}[t]
    
    \centering
    \scalebox{0.8}{
    \begin{tabular}{l l r@{\hspace{0.15cm}}}
        \toprule
        \textbf{Hyper-parameter} & \textbf{Method} & \textbf{Value} \\
        \midrule
        \multirow{2}{*}{Batch size} & Finetuning                       & 64    \\
                            & Others                           & 32    \\
        \midrule          
        \multirow{2}{*}{Base LR}    & Rep., Frozen, Finetuning & 1E-03 \\
                            & Others                           & 1E-04 \\
        \midrule
        \multirow{3}{*}{Step}       & Rep.                    & 6     \\
                            & Frozen                           & 5     \\
                            & Others                           & -     \\
        \midrule
        BERT LR                     & \multirow{3}{*}{Finetuning}      & 1E-04 \\
        Warm-up Rate                &                                  & 0.1   \\
        \# Epochs                   &                                  & 15    \\
        \midrule
        Gamma                       & \multirow{3}{*}{All}             & 1E-02 \\
        Weight Decay                &                                  & 1E-05 \\
                
        \bottomrule
    \end{tabular}
    }
    \caption{Hyper-parameters used to train ExCL with BERT on the Charades-STA dataset.}
    \label{table:excl_BERT_Charades_param}
\end{table}

    


\begin{table}[t]
    
    \centering
    \scalebox{0.8}{
    \begin{tabular}{l l r@{\hspace{0.15cm}}}
        \toprule
        \textbf{Hyper-parameter} & \textbf{Method} & \textbf{Value} \\
        \midrule
        \multirow{2}{*}{Batch size}   & Reproduction                          & 32    \\
                                      & Others                                & 64    \\
                                      \midrule
        \multirow{2}{*}{Base LR}      & \thead{ \Pfeiffer, \Parallel, \\ \Compacter, \Prefix}  & 1E-04 \\
                                      & Others                                & 1E-03 \\
                                      \midrule
        \multirow{2}{*}{Step}         & Finetuning                            & -     \\
                                      & Others                                & 5     \\
                                      \midrule
        \multirow{2}{*}{Weight Decay} & \thead{ \Pfeiffer, \Parallel, \\ \Compacter, \Prefix}  & 1E-04 \\
                                      & Others                                & 1E-06 \\
                                      \midrule
        BERT LR                       & \multirow{3}{*}{Finetuning}           & 1E-04 \\
        Warm-up Rate                  &                                       & 0.2   \\
        \# Epochs                     &                                       & 15    \\
        \midrule
        Gamma                         & All                                   & 1E-02\\
                
        \bottomrule
    \end{tabular}
    }
    \caption{Hyper-parameters used to train ExCL with BERT on the ActivityNet dataset.}
    \label{table:excl_BERT_Anet_params}
\end{table}



    
    
\begin{table}[t]
    
    \centering
    \scalebox{0.8}{
    \begin{tabular}{l l r@{\hspace{0.15cm}}}
        \toprule
        \textbf{Hyper-parameter} & \textbf{Method}  & \textbf{Value} \\
        \midrule
        Batch size                    & \multirow{3}{*}{All}        & 32    \\
        Base LR                       &                             & 1E-03 \\
        Gamma                         &                             & 1E-02 \\
        \midrule
        \multirow{2}{*}{Step}         & LoRa                        & 8     \\
                                      & Others                      & -     \\
                                      \midrule
        \multirow{2}{*}{Weight Decay} & Finetunning, Frozen         & 1E-05 \\
                                      & Others                      & 1E-04 \\
                                      \midrule
        BERT LR                       & \multirow{3}{*}{Finetuning} & 1E-04 \\
        Warm-up Rate                  &                             & 0.2   \\
        \# Epochs                     &                             & 15    \\
        
        \bottomrule
    \end{tabular}
    }
    \caption{Hyper-parameters used to train ExCL with BERT on the YouCookII dataset.}
    \label{table:excl_BERT_Youcookii_params}
\end{table}




\begin{table}[t]
    
    \centering
    \scalebox{0.8}{
    \begin{tabular}{l l r@{\hspace{0.15cm}}}
        \toprule
        \textbf{Hyper-parameter} & \textbf{Method} & \textbf{Value} \\
        \midrule
        Batch size            & \multirow{4}{*}{All}        & 256   \\
        Base LR               &                             & 1E-04 \\
        Weight Decay          &                             & 1E-05 \\
        Gamma                 &                             & 1E-02 \\
        \midrule
        \multirow{2}{*}{Step} & Finetuning                  & -     \\
                              & Others                      & 6     \\
        \midrule
        BERT LR               & \multirow{3}{*}{Finetuning} & 1E-04 \\
        Warm-up Rate          &                             & 0.1   \\
        \# Epochs             &                             & 20   \\
             
        \bottomrule
    \end{tabular}
    }
    \caption{Hyper-parameters used to train TMLGA with BERT on the Charades-STA dataset.}
    \label{table:tmlga_BERT_Charades_param}
\end{table}



\begin{table}[t]
    
    \centering
    \scalebox{0.8}{
    \begin{tabular}{l l r@{\hspace{0.15cm}}}
        \toprule
        \textbf{Hyper-parameter} & \textbf{Method} & \textbf{Value} \\
        \midrule
        Batch size            & \multirow{4}{*}{All}        & 64   \\
        Base LR               &                             & 1E-04 \\
        Weight Decay          &                             & 1E-05 \\
        Gamma                 &                             & 1E-02 \\
        \midrule
        \multirow{2}{*}{Step} & Finetuning                  & -     \\
                              & Others                      & 5     \\
        \midrule
        BERT LR               & \multirow{3}{*}{Finetuning} & 1E-04 \\
        Warm-up Rate          &                             & 0.1   \\
        \# Epochs             &                             & 15   \\   
        \bottomrule
    \end{tabular}
    }
    \caption{Hyper-parameters used to train TMLGA with BERT on the ActivityNet dataset. }
    \label{table:tmlga_BERT_Anet_param}
\end{table}



\begin{table}[t]
    
    \centering
    \scalebox{0.8}{
    \begin{tabular}{l l r@{\hspace{0.15cm}}}
        \toprule
        \textbf{Hyper-parameter} & \textbf{Method} & \textbf{Value} \\
        \midrule
        Batch size                    & \multirow{4}{*}{All}        & 64    \\
        Base LR                       &                             & 1E-03 \\
        Step                          &                             & 6     \\
        Gamma                         &                             & 1E-02 \\
        \midrule
        \multirow{2}{*}{Weight Decay} & Prefix                      & 1E-04 \\
                                      & Others                      & 1E-05 \\
        \midrule
        BERT LR                       & \multirow{3}{*}{Finetuning} & 1E-04 \\
        Warm-up Rate                  &                             & 0.2   \\
        \# Epochs                     &                             & 15  \\   
        \bottomrule
    \end{tabular}
    }
    \caption{Hyper-parameters used to train TMLGA with BERT on the YouCookII dataset. }
    \label{table:tmlga_BERT_YouCookII_param}
\end{table}




\begin{table}[t]
    
    \centering
    \scalebox{0.8}{
    \begin{tabular}{l l r@{\hspace{0.15cm}}}
        \toprule
        \textbf{Hyper-parameter} & \textbf{Method} & \textbf{Value} \\
        \midrule
        Batch size                    & \multirow{3}{*}{All}    & 5     \\
        Base LR                       &                         & 1E-04 \\
        Gamma                         &                         & 1E-02 \\
        \midrule
        \multirow{2}{*}{Weight Decay} & Frozen                  & 1E-04 \\
                                      & Others                  & 1E-05 \\
        \midrule
        \multirow{4}{*}{Step}         & \Prefix                  & 3     \\
                                      & \Houlsby, \Pfeiffer, \Lora & 4     \\
                                      & \Inverse, \Compacter      & 5     \\
                                      & Frozen                  & 6   \\   
        \bottomrule
    \end{tabular}
    }
    \caption{Hyper-parameters used to train DORi with BERT on the Charades-STA dataset. }
    \label{table:dori_BERT_Charades_param}
\end{table}


\begin{table}[t]
    
    \centering
    \scalebox{0.8}{
    \begin{tabular}{l l r@{\hspace{0.15cm}}}
        \toprule
        \textbf{Hyper-parameter} & \textbf{Method} & \textbf{Value} \\
        \midrule
        \multirow{2}{*}{Batch Size}   & Rep., Frozen         & 8     \\
                                      & Others               & 4     \\
        \midrule
        Base LR                       & \multirow{2}{*}{All} & 1E-04 \\
        Gamma                         &                      & 1E-02 \\
        \midrule
        \multirow{2}{*}{Weight Decay} & Rep., Frozen         & 1E-04 \\
                                      & Others               & 1E-05 \\
        \midrule
        \multirow{3}{*}{Step}         & Rep.                 & 3     \\
                                      & All adapters         & 4     \\
                                      & Frozen               & 6   \\   
        \bottomrule
    \end{tabular}
    }
    \caption{Hyper-parameters used to train DORi with BERT on the ActivityNet Captions dataset. }
    \label{table:dori_BERT_Anet_param}
\end{table}




\begin{table}[t]
    
    \centering
    \scalebox{0.8}{
    \begin{tabular}{l l r@{\hspace{0.15cm}}}
        \toprule
        \textbf{Hyper-parameter} & \textbf{Method} & \textbf{Value} \\
        \midrule
        \multirow{2}{*}{Batch Size}   & Rep.                 & 2     \\
                                      & Others               & 4     \\
        \midrule
        Base LR                       & \multirow{2}{*}{All} & 1E-04 \\
        Step                          &                      & 6     \\
        \midrule
        \multirow{2}{*}{Weight Decay} & Rep.                 & 1E-05 \\
                                      & Others               & 1E-04   \\   
        \midrule
        \multirow{2}{*}{Gamma}        & Rep.                 & 1E-02 \\
                                      & Others               & 1E-01   \\
        \bottomrule
    \end{tabular}
    }
    \caption{Hyper-parameters used to train DORi with BERT on the YouCookII dataset. }
    \label{table:dori_BERT_YouCookII_param}
\end{table}

\begin{table*}[t]
    
    \centering
    \scalebox{0.8}{
    \begin{tabular}{l@{\hspace{0.15cm}} c@{\hspace{0.15cm}} c@{\hspace{0.15cm}} c@{\hspace{0.15cm}} c@{\hspace{0.15cm}} c@{\hspace{0.15cm}} c@{\hspace{0.15cm}} c@{\hspace{0.15cm}} c@{\hspace{0.15cm}}}
        \toprule
        \multirow{2}{*}{\bf Model} & \multicolumn{4}{c}{\bf Charades-STA} & \multicolumn{4}{c}{\bf ActivityNet} \\ 
        \cmidrule{2-9}
        & R@0.3 & R@0.5 & R@0.7 & mIoU & R@0.3 & R@0.5 & R@0.7 & mIoU\\
        \midrule
        TMLGA (orig.)       & 67.53 & 52.02 & 33.74 & 48.22 & 51.28 & 33.04 & 19.26 & 37.78  \\
        TMLGA (ours)        & 69.49 & 49.97 & 32.72 & 48.29 & 50.84 & 31.13 & 17.86 & 36.90 \\
        \midrule
        + BERT \Snowflake    & 70.08 & 49.92 & 31.42 & 48.34 & 52.10 & 32.57 & 18.64 & 37.63 \\ 
        \, + \Pfeiffer      & \und{71.64} & 51.59 & 33.41 & 49.50 & \und{53.98} & \und{35.20} & \und{20.43} & \und{38.88}  \\
        \, + \Houlsby       & 70.97 & 51.69 & 33.84 & 49.31 & \und{53.98} & 34.13 &  19.48 & 38.57 \\
        \, + \Prefix        & 71.40 & 52.53 & 33.82 & 49.57 & 53.61 & 34.03 & 19.89 & 38.34  \\
        \, + \Inverse       & 71.02 & \und{52.77} & \und{34.49} & 49.33 & 52.47 & 33.76 & 19.93 & 37.93 \\
        \, + \Compacter     & 70.27 & 49.73 & 31.37 & 48.31 & 52.15 & 33.85 & 19.62 & 37.78 \\
        \, + \Lora          & 70.94 & 50.24 & 32.15 & 48.81 & 51.90 & 32.81 & 18.77 & 37.37 \\
        \, + \Parallel     & 70.48 & 51.64 & 33.66 & 49.33 & 43.50 & 23.20 & 11.59 & 32.29 \\
        \, + Fine-tuning    & 71.02 & 52.53 & 33.52 & \und{49.80} & 53.59 & 34.05 & 19.51 & 37.92 \\
        \midrule
        + RoBERTa \Snowflake  & 69.73 & 51.34 & 33.49 & 48.91 & 52.58 & 33.8  & 19.62 & 37.89 \\
        \, + \Pfeiffer      & 71.72 & \textbf{\und{53.84}} & \textbf{\und{34.78}} & \textbf{\und{49.91}} & \underline{54.51} & \underline{35.27} & 20.26 & 38.77 \\
        \, + \Houlsby       & 71.08 & 52.98 & 34.19 & 49.28 & 53.56 & 34.34 & 20.16 & 38.89 \\
        \, + \Prefix        & \textbf{\und{72.28}} & 52.42 & 33.98 & 49.90 & 53.08 & 33.19 & 19.48 & 38.47 \\
        \, + \Inverse       & 71.53 & 52.69 & 33.98 & 49.50 & 54.36 & 34.85 & \underline{20.46} & \underline{39.35} \\
        \, + \Compacter     & 71.21 & 51.56 & 33.17 & 49.20 & 52.88 & 33.16 & 19.35 & 38.09 \\
        \, + \Lora          & 71.64 & 51.88 & 33.17 & 49.55 & 53.73 & 34.00 & 19.43 & 38.64 \\
        \, + \Parallel     & 70.99 & 52.93 & 34.01 & 49.19 & 43.50 & 23.20 & 11.59 & 32.29 \\
        \, + Fine-tuning    & 71.61 & 53.15 & 33.33 & 49.77 & 52.70 & 33.21 & 19.69 & 38.47	\\
        \midrule
        + DeBERTa \Snowflake & 70.73 & 52.53 & 33.49 & 49.32 & 53.04 & 33.94 & 20.22 & 38.72 \\
        \, + \Pfeiffer      & 71.34 & \und{53.49} & 34.65 & \und{49.78} & 54.37 & 34.70 & 20.49 & 39.30\\
        \, + \Houlsby       & 70.83 & 52.10 & 33.74 & 49.48 & 53.25 & 33.93 & 19.55 & 38.45 \\
        \, + \Inverse       & \und{71.64} & 52.58 & 33.95 & 49.66 & \textbf{\und{55.09}} & \textbf{\und{35.45}} & \textbf{\und{20.66}} & \textbf{\und{39.71} }\\
        \, + \Compacter     & 70.08 & 51.64 & 32.98 & 48.78 & 53.60 & 34.21 & 20.14 & 38.79 \\
        \, + \Lora          & 71.18 & 52.34 & 33.31 & 49.18 & 53.98 & 34.53 & 20.02 & 39.00 \\
        \, + \Parallel*     & 70.73 & 53.36 & \und{34.76} & 49.35 & 43.50 & 23.20 & 11.59 & 32.29 \\
        \, + Fine-tuning    & 71.59 & 53.44 & 33.44 & 49.63 & 53.54 & 33.78 & 20.12 & 38.92	\\
        \bottomrule
    \end{tabular}
    }
    \caption{Detailed results combining the TMLGA model with our three pre-trained language encoders and adapters, tested on Charades-STA and ActivityNet Captions. Underlined results indicate the best performance within the model and dataset combination, while the results in bold indicate the best performance within the dataset.}
    \label{table:TMLGA-ANet-Charades_full}
\end{table*}

\begin{table}[t]
    
    \centering
    \scalebox{0.8}{
    \begin{tabular}{l l r@{\hspace{0.15cm}}}
        \toprule
        \textbf{Hyper-parameter} & \textbf{Method} & \textbf{Value} \\
        \midrule
        \multirow{2}{*}{Batch Size} & Finetuning                  & 256   \\
                                    & Others                      & 64    \\
        \midrule
        Base LR                     & \multirow{4}{*}{All}        & 1E-04 \\
        Weight Decay                &                             & 1E-05 \\
        Gamma                       &                             & 1E-02 \\
        Step                        &                             & 6     \\
        \midrule
        RoBERTa LR                     & \multirow{3}{*}{Finetuning} & 1E-04 \\
        Warm-up Rate                &                             & 0.3   \\
        \# Epochs                   &                             & 10   \\
        \bottomrule
    \end{tabular}
    }
    \caption{Hyper-parameters used to train TMLGA with RoBERTa on the Charades-STA dataset. }
    \label{table:tmlga_roberta_Charades_param}
\end{table}



\begin{table}[t]
    
    \centering
    \scalebox{0.8}{
    \begin{tabular}{l l r@{\hspace{0.15cm}}}
        \toprule
        \textbf{Hyper-parameter} & \textbf{Method} & \textbf{Value} \\
        \midrule
        Batch Size                  & \multirow{5}{*}{All}        & 64    \\
        Base LR                     &                             & 1E-04 \\
        Weight Decay                &                             & 1E-05 \\
        Gamma                       &                             & 1E-02 \\
        Step                        &                             & 6     \\
        \midrule
        RoBERTa LR                     & \multirow{3}{*}{Finetuning} & 1E-04 \\
        Warm-up Rate                &                             & 0.2   \\
        \# Epochs                   &                             & 10      \\
        \bottomrule
    \end{tabular}
    }
    \caption{Hyper-parameters used to train TMLGA with RoBERTa on the ActivityNet Captions dataset. }
    \label{table:tmlga_roberta_Anet_param}
\end{table}





\begin{table}[t]
    
    \centering
    \scalebox{0.8}{
    \begin{tabular}{l l r@{\hspace{0.15cm}}}
        \toprule
        \textbf{Hyper-parameter} & \textbf{Method} & \textbf{Value} \\
        \midrule
        \multirow{2}{*}{Batch Size} & Finetuning                  & 256   \\
                                    & Others                      & 64    \\
        \midrule
        Base LR                     & \multirow{3}{*}{All}        & 1E-04 \\
        Weight Decay                &                             & 1E-05 \\
        Gamma                       &                             & 1E-02 \\
        \midrule
        Step                        & \Houlsby                    & 4     \\
                                    & \Inverse                    & 5     \\
                                    &  \thead{ Frozen, \Pfeiffer, \\ \Parallel, \Lora } & 6     \\
                                    & \Compacter                   & 7     \\
        \midrule
        BERT LR                     & \multirow{3}{*}{Finetuning} & 1E-04 \\
        Warm-up Rate                &                             & 0.3   \\
        \# Epochs                   &                             & 10       \\
        \bottomrule
    \end{tabular}
    }
    \caption{Hyper-parameters used to train TMLGA with DeBERTa on the Charades-STA dataset. }
    \label{table:tmlga_deberta_charades_param}
\end{table}




\begin{table}[t]
    
    \centering
    \scalebox{0.8}{
    \begin{tabular}{l l r@{\hspace{0.15cm}}}
        \toprule
        \textbf{Hyper-parameter} & \textbf{Method} & \textbf{Value} \\
        \midrule
        Batch Size   & \multirow{4}{*}{All}        & 64    \\
        Base LR      &                             & 1E-04 \\
        Weight Decay &                             & 1E-05 \\
        Gamma        &                             & 1E-02 \\
        \midrule
        Step         & \Inverse                    & 5     \\
                     & Others                      & 6     \\
        \midrule
        BERT LR      & \multirow{3}{*}{Finetuning} & 1E-04 \\
        Warm-up Rate &                             & 0.2   \\
        \# Epochs    &                             & 10       \\
        \bottomrule
    \end{tabular}
    }
    \caption{Hyper-parameters used to train TMLGA with DeBERTa on the ActivityNet Captions dataset. }
    \label{table:tmlga_deberta_anet_param}
\end{table}

\begin{table}[t]
    
    \centering
    \scalebox{0.8}{
    \begin{tabular}{l l l r@{\hspace{0.15cm}}}
        \toprule
        Dataset                        & Method                     & Hyper-parameter & Value \\
        \midrule
                                       &                            & Batch Size      & 5     \\
                                       &                            & Base LR         & 1E-04 \\
                                       &                            & Weight Decay    & 1E-04 \\
                                       &                            & Gamma           & 5     \\
        \multirow{-5}{*}{Charades-STA} & \multirow{-5}{*}{Pfeiffer} & Step            & 1E-02 \\
        \midrule
                                       &                            & Batch Size      & 4     \\
                                       &                            & Base LR         & 1E-04 \\
                                       &                            & Weight Decay    & 1E-05 \\
                                       &                            & Gamma           & 4     \\
        \multirow{-5}{*}{ANet}         & \multirow{-5}{*}{Inverse}  & Step            & 1E-02\\
        \bottomrule
    \end{tabular}
    }
    \caption{Hyper-parameters used to obtain the best results for DORi with DeBERTa on the Charades-STA and ActivityNet Captions  datasets. }
    \label{table:dori_deberta}
\end{table}

\end{document}